\documentclass[letterpaper, 11pt]{article}

\usepackage{kotex}

\usepackage[margin=1in]{geometry}
\usepackage{setspace}
\usepackage{microtype}

\usepackage{amsmath} \allowdisplaybreaks
\usepackage{algorithm}
\usepackage[noend]{algpseudocode}
\usepackage{amssymb}
\usepackage{amsthm} \theoremstyle{definition}
\usepackage{esvect}

\newtheorem{lemma}{Lemma}

\newtheorem{proposition}{Proposition}
\newtheorem{definition}{Definition}

\usepackage{natbib}
\usepackage{bibentry}

\usepackage{multirow}
\usepackage{pdflscape}
\usepackage{array}
\usepackage{booktabs}

\newcolumntype{R}[1]{>{\raggedleft\arraybackslash}m{#1}}  %
\newcolumntype{C}[1]{>{\centering\arraybackslash}m{#1}}   %

\usepackage{graphicx}
\usepackage{svg}
\usepackage{caption}
\usepackage{subcaption}
\usepackage{appendix}

\usepackage[hyphens]{url}
\usepackage[bookmarks=false,hidelinks]{hyperref}

\usepackage{tikz}
\usepackage{pgfplots}
\usetikzlibrary{positioning}
\usetikzlibrary{arrows,shapes}
\usetikzlibrary{shapes.geometric}

\usepackage{authblk}

\usepackage[shortlabels]{enumitem}

\newcommand{\email}[1]{{\href{mailto:#1}{\nolinkurl{#1}}}}

\usepackage{bm}
\renewcommand{\vec}[1]{\boldsymbol{\mathbf{#1}}}
\newcommand{\mat}[1]{\vec{#1}}

\usepackage{xcolor}

\usepackage{etoolbox}
\makeatletter
\def\do#1{\@namedef{#1c}{\ensuremath{\mathcal{#1}}}}
\docsvlist{A,B,C,D,E,F,G,H,I,J,K,L,M,N,O,P,Q,R,S,T,U,V,W,X,Y,Z}
\makeatother

\renewcommand{\hat}{\widehat}

\DeclareMathOperator*{\argmax}{\arg \max}

\title{The Iterative Chainlet Partitioning Algorithm for the Traveling Salesman Problem with Drone and Neural Acceleration}

\author[1]{Jae Hyeok Lee}
\author[1]{Minwoo Kim}
\author[2]{Minjun Kim}
\author[1,3]{Jinkyoo Park}
\author[1,3]{Changhyun Kwon\thanks{Corresponding author}}

\affil[1]{Department of Industrial and Systems Engineering, KAIST, Daejeon, 34141, Republic of Korea}
\affil[2]{Samsung Advanced Institute of Technology, Suwon-si, Gyeonggi-do, 16678, Republic of Korea}
\affil[3]{Omelet, Inc., Daejeon, 34051, Republic of Korea}

\date{}

\begin{document}
\maketitle

\begin{abstract}
This study introduces the Iterative Chainlet Partitioning (ICP) algorithm and its neural acceleration for solving the Traveling Salesman Problem with Drone (TSP-D). The proposed ICP algorithm decomposes a TSP-D solution into smaller segments called chainlets, each optimized individually by a dynamic programming subroutine. The chainlet with the highest improvement is updated, and the procedure is repeated until no further improvement is possible. We show that the subroutine runs in quadratic time and the number of subroutine calls is bounded linearly in problem size for the first iteration and remains constant in subsequent iterations, ensuring algorithmic scalability. Empirical results show that ICP outperforms existing algorithms in both solution quality and computational time. Tested over 1,249 benchmark instances, ICP yields an average improvement of 2.6\% in solution quality over the previous state-of-the-art algorithm while reducing computational time by 91.3\%. The procedure is deterministic, ensuring reliability without requiring multiple runs. The subroutine is the computational bottleneck in the already efficient ICP algorithm. To reduce the necessity of subroutine calls, we integrate a graph neural network (GNN) to predict incremental improvements. We demonstrate that the resulting Neuro ICP (NICP) achieves substantial acceleration while maintaining solution quality. Compared to ICP, NICP reduces the total computational time by 28.6\%, while the objective function value increase is limited to 0.14\%. A transfer learning framework enables efficient extension to various operational constraints, making this a valuable foundation for developing efficient algorithms for truck-drone synchronized routing problems.
\paragraph{Keywords:} traveling salesman problem; drones; deep learning; cost prediction; transfer learning
\end{abstract}

\section{Introduction}
\label{sec:introduction}

The growing demand for faster delivery challenges traditional logistics, leading to the exploration of drones as a promising alternative \citep{pwc2024drone}. 
However, the short flight ranges of drones make it unlikely they can be used as standalone delivery vehicles \citep{jazairy2025drones}. 
This has led to a focus on \emph{drone-assisted delivery}, where trucks are responsible for some customer demands while drones launch from and land on them to complete other deliveries, thereby preserving battery life. 
The significance of this research extends beyond truck-drone coordination. 
The problem provides a basis for developing methods for coordinated operations between heterogeneous entities---such as autonomous robots and a human workforce---performing complementary logistics tasks like deliveries and warehouse operations \citep{keith2024review}.

Truck-drone combined routing problems vary widely, encompassing different fleet compositions, operational rules, and optimization objectives. 
Within this broad domain, our work focuses on the synchronized operation of a truck and a drone serving customers while minimizing completion time. 
For comprehensive reviews of other problem configurations in truck-drone routing, we refer readers to surveys: 
\cite{macrina2020drone}, \cite{luo2025mathematical} and \cite{zhou2025survey}.

The Flying Sidekick Traveling Salesman Problem (FS-TSP) first formalized single truck-drone synchronized delivery \citep{murray2015flying}.
The problem introduces the concept of a \textit{sortie}, where a unit-capacity drone launches from the truck at one customer location, serves an eligible customer and returns to the truck at a different customer location.
Each sortie requires fixed service times for launch and rendezvous, and the drone's flight duration is constrained by battery life until rendezvous with the truck. 
The problem prohibits revisiting customer locations, and a drone cannot be redeployed once it returns to the depot. 
The objective is to minimize the total time required to serve all customers and return both vehicles to the depot.

In contrast, the Traveling Salesman Problem with Drone (TSP-D), a variant of the FS-TSP, captures the essence of the problem's collaborative nature among heterogeneous vehicles \citep{agatz2018optimization}. 
The TSP-D defines each drone delivery as an \textit{operation}—analogous to sortie of FS-TSP—where a drone departs from the truck to serve a customer before returning for rendezvous. 
The TSP-D differs from FS-TSP by eliminating all drone-related service times and allowing more flexible operations:
Trucks may revisit customer locations, and drones may execute \emph{loops} by returning to their departure location.
Additionally, drones may land while awaiting truck rendezvous.
Various operational constraints exist; the variant without truck revisitation, loops, customer eligibility restrictions, and drone flying range limitations is called the basic TSP-D \citep{roberti2021exact, blufstein2024decremental}, which we will refer to simply as TSP-D throughout this paper.

\begin{figure}
    \centering
    \includegraphics[width=0.95\textwidth]{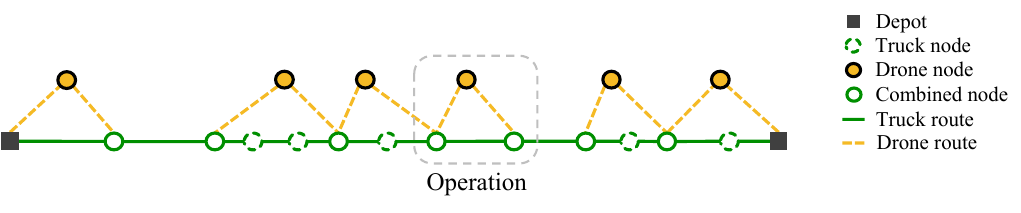}
    \caption{Illustration of Node Types in TSP-D Solution}
    \label{fig:tspd_solution}
\end{figure}

The TSP-D is defined on a complete graph \( \Gc \) with node set \( \Nc = \{ 1, 2, \cdots, N\}\), where node 1 represents the depot. 
The objective is to determine synchronized truck and drone routes that serve all \( N - 1\) customer nodes exactly once and return to the depot in a minimal time, where the drone operates at a relative speed \(\alpha\) to the truck.
The drone operates under several assumptions:
\begin{enumerate}
        \item The drone has a unit capacity requiring return to the truck after each delivery;
        \item The drone has unlimited flying range and can visit any customer location;
    \item Parcel pickup and drone landings are restricted to customer locations or the depot;
    \item Loops are prohibited. That is, launch and landing locations are required to be distinct; and
    \item All drone-related service times are negligible.
\end{enumerate}

Figure~\ref{fig:tspd_solution} illustrates a TSP-D solution where nodes serve three distinct roles: \textit{truck nodes} visited exclusively by the truck, \textit{drone nodes} serviced solely by the drone, and \textit{combined nodes} serving as synchronization points for both vehicles. 
Each \textit{operation} consists of exactly two combined nodes (start and end), at most one drone node, and zero or more truck nodes.

Despite simplification, coordinating the truck and drone on time remains challenging.
Its inherent NP-hardness poses significant scalability challenges for exact algorithms, driving the need for heuristic solution methods.
Researchers have explored a variety of approaches to tackle these challenges, ranging from dynamic programming \citep{agatz2018optimization} to advanced metaheuristic algorithms \citep{de2020variable, mahmoudinazlou2024hybrid} and machine learning-based strategies \citep{bogyrbayeva2023deep}. 
While previous approaches have significantly progressed, challenges remain in balancing solution quality and computational efficiency.

In this paper, we aim to develop an approach that enhances both the quality and efficiency of solutions, addressing the trade-offs that have limited the performance of existing methods.
We make three main methodological contributions. 
\emph{First}, we re-examine the dynamic programming subroutine at the core of many TSP-D heuristics. 
By formalizing its pruning strategy, we show that the Exact Partitioning (EP) algorithm in fact runs in $O(N^2)$, in the worst case.
We further generalize this result, incorporating additional operational constraints.
\emph{Second}, leveraging this accelerated subroutine, we propose the Iterative Chainlet Partitioning (ICP) algorithm for the TSP-D that finds solutions superior to the best-known solutions so far for many instances while reducing computational time significantly.
In ICP, the TSP-D route is divided into smaller segments called \emph{chainlets}, which are iteratively optimized by a precise subroutine powered by our enhanced partitioning algorithm.
The scalability of this approach is supported by our theoretical analysis, which derives a closed-form upper bound on the number of required subroutine calls.
\emph{Third}, we implement a machine-learning method to enhance the algorithm's computational efficiency further. 
Specifically, we employ a graph neural network (GNN) that predicts the potential improvement of each chainlet. 
This reduces the need for direct executions of subroutines, accelerating the process while maintaining solution quality.

Our proposed framework offers several distinctive advantages. 
First, unlike many high-quality solution methods, the ICP algorithm itself is fully \emph{deterministic}, guaranteeing consistent, high-quality solutions in a single run. 
The framework is also \emph{flexible} and \emph{adaptable}, supporting key operational constraints and making it a strong foundation for a wide range of truck-drone routing problems. 
We demonstrate that ICP is also highly effective for the TSP-D with flying range limitations, a key variant for real-world applications.
Second, the neural acceleration framework is designed to be highly practical. 
By constraining the GNN to small, fixed-size inputs, our model is lightweight, trains quickly, and avoids common generalization challenges, resulting in robust performance on out-of-distribution (OOD) data without retraining.
Finally, our novel transfer learning strategy extends this practicality; by integrating a crafted Feature-wise Linear Modulation (FiLM) layer \citep{perez2018film}, the model can adapt to new operational constraints via lightweight training while preventing catastrophic forgetting \citep{kirkpatrick2017overcoming}.

The rest of this paper is organized as follows.
Section~\ref{sec:literature_review} briefly summarizes the related literature.
Section~\ref{sec:ep} presents the Exact Partitioning and its analysis.
Section~\ref{sec:icp} proposes the ICP algorithm.
Section~\ref{sec:nicp} details the neural acceleration framework.
The performance of our methods is tested through extensive numerical experiments in Section~\ref{sec:computational_study}, and we conclude the paper in Section~\ref{sec:conclusion}.

\section{Literature Review}
\label{sec:literature_review}

This section positions our work by reviewing solution methodologies for the TSP-D, comparing ICP to the decomposition-based metaheuristics for the large-scale routing problem, and contextualizing our neural acceleration within machine learning for combinatorial optimization.

\subsection{Solution Methodologies for TSP-D}
\label{sec:solution_method_tspd}

We begin by examining the capabilities of exact methods.
\cite{roberti2021exact} proposed a compact mixed-integer linear program that synchronizes truck and drone flows to solve instances with up to 14 customers, while their more branch-and-price algorithm, using dynamic programming to solve an \textit{ng}-route relaxation, solved instances up to 29 customers within an hour.
Building upon this, \cite{blufstein2024decremental} introduced several enhancements, including tailored decremental state-space relaxation and variable fixing, extending solvability to instances with up to 59 customers---the current state-of-the-art for exact methods, to the best of our knowledge, regardless of the TSP-D variant.
For a comprehensive overview of other exact approaches, we refer the reader to the detailed review and summary table provided by \cite{blufstein2024decremental}.

Given these computational challenges, several heuristic approaches have been developed. 
\cite{agatz2018optimization} introduced an exact partitioning algorithm, \textsc{TSP-EP}, that uses dynamic programming to find a minimal TSP-D solution where truck and drone routes are subsequences of an initial TSP route.
They extended \textsc{TSP-EP} to \textsc{TSP-EP-all} by incorporating local search methods sequentially to perturb the initial TSP route. 
Although efficient and effective for instances with fewer than 20 customers \citep{bogyrbayeva2023deep}, each iteration of \textsc{TSP-EP-all} incurs a computational cost of $O(N^5)$—combining $O(N^3)$ dynamic programming and $O(N^2)$ local search—limits its scalability, making it more valuable as a subroutine.

Building on this strength, inspired by the divide-and-conquer heuristic (DCH) \citep{poikonen2019branch}, \cite{bogyrbayeva2023deep} proposed the Divide-Partition-and-Search (\( \textsc{DPS}_{25} \)), partitioning the problem into subgroups containing 25 nodes each and applying \textsc{TSP-EP-all} to optimize individual subgroups.
It demonstrates moderate solution quality while maintaining computational efficiency for large-scale instances.
Additionally, they developed an end-to-end deep reinforcement learning model, \( \text{HM}_{4800} \). 
Leveraging graphics processing unit (GPU) parallelism, it efficiently and effectively solves large instances but remains restricted to its specific training configurations.
A variety of metaheuristic approaches have also been proposed, including Greedy Randomized Adaptive Search Procedure (GRASP) \citep{ha2018min} and Hybrid General Variable Neighborhood Search with a tabu search phase (HTGVNS) \citep{freitas2023exact}, while Hybrid Genetic Algorithm with type-aware chromosomes (HGA-TAC$^+$) \citep{mahmoudinazlou2024hybrid} is considered the state-of-the-art, achieving the best-known solution quality on standard benchmarks for the TSP-D and its variants. 
Although it is originally described for FS-TSP and TSP-D with loops, computational results are reported for the TSP-D.

While existing approaches have improved TSP-D solution capabilities, they face inherent limitations. 
Metaheuristic approaches deliver high-quality solutions, but their probabilistic nature requires substantial computational time and multiple runs, which introduces variability in solution quality. 
An end-to-end learning method offers computational efficiency but is limited to problems that match its training parameters and still inherits stochastic characteristics. 

Our proposed ICP algorithm addresses these limitations by combining deterministic procedures with high solution quality and computational efficiency. 
Unlike existing methods, ICP requires no multiple trials to achieve consistent results and functions effectively across various problem configurations, even with neural acceleration. 
This positions ICP as a significant advancement in solving TSP-D instances, particularly for large-scale applications.

\subsection{Comparison with POPMUSIC}
\label{sec:popmusic}

\begin{figure}
    \centering
    \includegraphics[width=\textwidth]{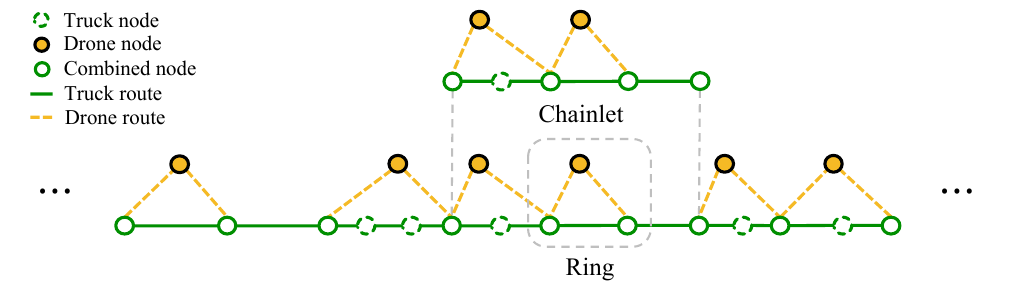}
    \caption{Conceptual Framework for Understanding TSP-D Tour as a Chain}
    \label{fig:chainlet_concept}
\end{figure}

ICP shares similarities with the Partial OPtimization Metaheuristic Under Special Intensification Conditions (POPMUSIC) \citep{ribeiro2002popmusic}, a general method for large-scale combinatorial optimization successfully applied to various routing problems.
The framework operates by decomposing a solution and iteratively finding compatible improvements within subproblems, ensuring local gains translate directly into global enhancements.
Adapting this template to the TSP-D requires synchronizing the truck and drone to maintain feasibility in partial solutions. 
We address this with our concept of a \emph{chainlet}: the tour is a \emph{chain} of operations, or \emph{rings}, and a chainlet is an optimizable segment of this chain (Figure~\ref{fig:chainlet_concept}). 
Because our TSP-ep-all subroutine preserves the start and end nodes of a chainlet, with careful initial tour construction, it can be optimized independently, guaranteeing that any local improvement enhances the overall solution.

ICP also diverges from standard POPMUSIC implementations in its selection strategy. 
Whereas most variants act as gradient methods that randomly or sequentially select subproblems, ICP employs a steepest descent approach, greedily selecting the single chainlet that offers the maximum improvement in each iteration.
This greedy selection strategy provides two key advantages. 
 First, it makes the algorithm deterministic, ensuring stable performance. 
 Second, it provides a theoretical basis for scalability, a property merely observed empirically in prior work \citep{taillard2019popmusic}. 
 By optimizing only the most promising chainlet with a fixed subproblem size and caching results, we derive an upper bound on the number of subroutine calls that is linear in $N$ for the first iteration and constant thereafter. 
This rigorously grounds ICP's efficiency for large-scale instances.

\subsection{Learning Approaches for Routing Problems}

Solving combinatorial optimization (CO) with deep learning, known as Neural Combinatorial Optimization (NCO), has grown rapidly.
As the literature is now extensive, we briefly introduce the main paradigms for routing problems and refer the interested reader to several recent surveys. 
For a comprehensive view of machine-learning approaches to routing problems, see \citet{zhou2025learning};
for GNN foundations in CO, see \citet{cappart2023combinatorial}; 
and for tutorial-style guidance in an operational research venue, see \citet{angioni2025neural}.

Early developments for routing included supervised learning and deep reinforcement learning approaches that construct solutions end-to-end \citep{bogyrbayeva2024machine}.
While such approaches have demonstrated that neural networks could learn heuristics for solving NP-hard combinatorial optimization problems and produce quality solutions quickly, a well-crafted, human-developed algorithm often outperforms by a large margin.
For instance, the Lin–Kernighan–Helsgaun (LKH) algorithm \citep{helsgaun2017extension} is reported to outperform end-to-end neural methods in solution quality and, in some cases, computational time on randomly generated TSP instances \citep{sui2025survey}.

Consequently, a new strand of \emph{hybrid} methodologies has emerged to bridge the gap between purely handcrafted methods and end-to-end learned approaches.
These hybrid methods either use machine learning to enhance classical algorithms by addressing specific subproblems or utilize traditional heuristics to improve an initial solution constructed by a learning model \citep{bogyrbayeva2024machine}.
In the context of truck-drone collaborative routing, hybrid strategies have been rarely explored:
\citet{boccia2021feature} and \citet{boccia2024new} employ supervised learning to classify customers suitable for drone delivery before solving the reduced formulation to optimality.
These hybrid methods have proven highly effective, maintaining the sophisticated structure of classical heuristics while strategically enhancing specific components through learning.

Our work reinforces this pattern: while the purely end-to-end learning-based approach, HM$_{4800}$, performs reasonably well, ICP, a carefully constructed heuristic, surpasses both solution quality and speed.
Therefore, we adopt a hybrid approach that accelerates ICP using a learning model to predict subproblem costs, similar to the strategy used for POPMUSIC in CVRP \citep{li2021learning}.
As numerical experiments confirm, our hybrid approach shows that integrating a learning model can accelerate an already efficient computational method without compromising solution quality.

Beyond this hybrid acceleration, our work presents a novel transfer learning strategy.
Transfer learning uses knowledge gained from one task to improve performance on a related task \citep{pan2009survey}, with pre-train-then-fine-tune as a canonical paradigm.
In NCO, this has been explored in several contexts, such as generalizing to larger problem scales \citep{zhou2024instance}, adapting to distribution shifts \citep{geisler2022generalization}, and transferring knowledge across different problem families \citep{lin2024cross}. 
Most efforts operate within end-to-end neural solvers; in hybrid NCO settings, transfer learning remains largely unexplored, with few exceptions \citep{correa2025tunensearch}.

We highlight the distinction of our transfer learning strategy in two key aspects.
First, to the best of our knowledge, this is the first work to leverage transfer learning within a neural surrogate predictor embedded in a classical algorithmic framework to extend capabilities to problem variants.
Second, we introduce a parameter-efficient fine-tuning strategy employing a FiLM layer designed to act as an identity function when the new constraint is inactive, thereby \emph{extending} rather than \emph{adapting} the original model as typically done in the NCO literature. 
This preserves the integrity of initially learned patterns and prevents catastrophic forgetting while enabling the model to handle new operational constraints with lightweight training.
Together, this demonstrates the extensibility of hybrid frameworks incorporating neural surrogate predictors to accommodate various variants, reinforcing the broader practicality of neural surrogate approaches.

\section{The Exact Partitioning Algorithm}
\label{sec:ep}

In this section, we introduce the Exact Partitioning (EP) algorithm of \cite{agatz2018optimization}.
We formalize its pruning strategy and prove that, in fact, EP admits a worst-case time complexity of $O(N^2)$.
This result remains valid under additional operational constraints, specifically customer eligibility and drone flying range limitations, requiring at most minor modifications.
We provide all proofs in Appendix~\ref{sec:proofs}.

We first present the formal definition of \emph{ring}, \emph{chainlet}, and \emph{chain}, and then classify different rings according to the number of drone and truck nodes they contain. 

\begin{figure}
   \centering
        \includegraphics[scale=1.0]{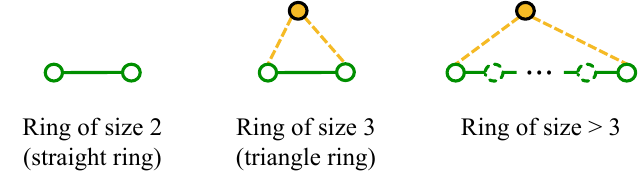}
    \caption{Illustration of Different Rings in a Chain}
    \label{fig:ring_types}
\end{figure}

\begin{definition}
A \emph{ring}, equivalent to an operation, consists of exactly two combined nodes serving as the start and end, at most one drone node, and zero or more truck nodes.
A \emph{chainlet} is a sequence of consecutive rings, and a \emph{chain} is the complete set of rings forming the TSP-D solution. 
The size of each structure is defined as the number of distinct nodes it contains.
\end{definition}

\begin{definition}
A ring of size 2 is called a \emph{straight ring}, and a ring of size greater than 2 is called a \emph{proper ring}.
In particular, a ring of size 3 is called a \emph{triangle ring}.
\end{definition}

Figure~\ref{fig:ring_types} illustrates the different ring types.
A straight ring does not utilize the drone and contains only two combined nodes.
A triangle ring employs the drone but has no intermediate truck nodes, meaning the truck moves directly to the rendezvous point.
As the effectiveness of deploying the drone decreases with lower $\alpha$, larger rings with additional truck nodes become more prevalent, reducing the total number of rings within a TSP-D chain.

The EP algorithm starts with a given TSP tour, represented as a sequence of nodes \(\Pc = (p_1, p_2, \ldots, p_{N+1})\), where \( p_1 = p_{N+1} = 1 \) is the depot.
A proper ring is represented by a triple of integers \((i,j,k)\), where \(i\), \(j\), and \(k\) denote 
the positions of the start, end, and drone nodes, respectively, within \(\Pc\). 
That is, node \(p_i\) is the start node, node \(p_j\) is the end node, and node \(p_k\) is the drone node. 
The set of proper rings is then defined as \(R = \{(i,j,k) \mid i < k < j\}\), obtained by assigning truck and drone nodes along the tour.
To incorporate operational constraints, we define \(\breve{R} \subseteq R\) as the set of feasible proper rings.

For each edge \((p_i, p_j)\), we denote the cost of the truck and the drone by \(c^t(p_i, p_j)\) and \(c^d(p_i, p_j)\), respectively.
We assume that \(c^t(p_i, p_j) = \alpha c^d(p_i, p_j)\) where \(\alpha \geq 1\) is consistent over the whole problem.
Define
\begin{equation}
S^t(i,j) = \sum_{l=i}^{j-1} c^t(p_l, p_{l+1}), \qquad S^t(i,i)=0,
\end{equation}
as the truck cost of traversing the subsequence \((p_i, p_{i+1}, \ldots, p_j)\).  
For each proper ring \((i,j,k)\), the truck and drone costs are given by
\begin{equation}
C^t(i,j,k) = S^t(i,k-1) + S^t(k+1,j) + c^t(p_{k-1}, p_{k+1}),
\end{equation}
\begin{equation}
C^d(i,j,k) = c^d(p_i, p_k) + c^d(p_k, p_j),
\end{equation}
and the ring cost is defined as
\begin{equation}
C(i,j,k) = \max \{ C^t(i,j,k), \, C^d(i,j,k) \}.
\end{equation}
For each consecutive subsequence \((p_i , p_{i+1}, \ldots, p_j)\) of \(\Pc\),
we compute the time duration of the best feasible ring---proper or straight---covering exactly this subsequence
\begin{equation}
T(i, j) = \begin{cases}
    \min_{k=i+1}^{j-1} C(i, j, k), & \text{ if } j > i+1\\
    c^t(p_i, p_j), & \text{ if } j = i+1
\end{cases}
\end{equation}
and the dynamic programming algorithm updates
\begin{equation} \label{value_function_update}
	V(0) = 0, \qquad 
	V(v) = \min_{u=0}^v [V(u) + T(u, v)].
\end{equation}
A naive implementation computes all \(T(i,j,k)\), of which there are \(O(N^3)\), and each evaluation requires \(O(N)\), yielding a total runtime of \(O(N^4)\). 
\cite{agatz2018optimization} observed that, exploiting simple recurrence relations, each \(T(i,j,k)\) can be updated in constant time, reducing the total complexity to the number of rings \(O(N^3)\). 
They further introduced a pruning strategy to discard dominated rings early, providing additional speedup.

We now show that leveraging the speedup, the number of candidate rings reduces to $O(N^2)$, yielding a worst-case complexity of $O(N^2)$ for the EP algorithm.
For each start node $i$ and drone node $k$, we first define the cutoff index
\begin{equation}
  E(i, k) := \min_{(i, j, k) \in \breve{R}} \{\, j : C^d(i, j, k) \leq C^t(i, j, k)\,\}.
\end{equation}
We distinguish between \emph{long} and \emph{short} rings: a proper ring $(i,j,k)$ is called \emph{long} if $j \leq E(i, k)$, and, otherwise, the ring is called \emph{short}.
The pruning strategy of \cite{agatz2018optimization} claims that long rings never contribute to the optimal values in \eqref{value_function_update},
and Lemma~\ref{lem:EP_long_ring_monotone} justifies the strategy.

\begin{lemma} \label{lem:EP_long_ring_monotone}
For any long ring $(i, j, k)$, $C^d(i, j, k) \leq C^t(i, j, k)$.
\end{lemma}

Let $\breve{R}^s$ denote the set of short feasible rings.
Lemma~\ref{lem:EP_num_short_ring} yields a quadratic bound on the number of short feasible rings.

\begin{lemma} \label{lem:EP_num_short_ring}
The total number of short feasible rings satisfies
\[
\big| \breve{R}^s \big| = \Big| \{(i, j, k) \in \breve{R} : j \leq E(i, k) \} \Big| < n^2.
\]
\end{lemma}

Lemmas~\ref{lem:EP_long_ring_monotone} and \ref{lem:EP_num_short_ring}, along with the constant-time recurrence for updating each $T(i,j,k)$, ensures that the EP algorithm for TSP-D operates in $O(N^2)$ time, looping only until the condition $C^d(i,j,k) \leq C^t(i,j,k)$ is reached.
Consequently, the complexity of a single iteration of the \textsc{TSP-ep-all} algorithm reduces from $O(N^5)$ to $O(N^4)$.

The result extends naturally to TSP-D variants with additional operational constraints:
\begin{itemize}
  \item \textbf{Customer eligibility restriction} Suppose only nodes in $\breve{N} \subseteq N$ can be selected as drone nodes. In this case, we restrict the outer loop over $k$ to those nodes, omitting ineligible drone nodes. The quadratic bound on the number of rings still applies, and the complexity of the EP algorithm remains $O(N^2)$.
  \item \textbf{Flying range limitation}  Suppose the drone has a limited flying range, which we define as a percentage, $f$, of the maximum inter-point distance in the instance, $d^{\max}$. 
 Since a drone operation involves two flight legs, a range of $f \ge 200$ percent is unlimited. 
 To enforce a finite range, we preprocess each node $u$ by sorting all potential rendezvous nodes in increasing order of drone distance $c^d(u,\cdot)$. 
 This sorting takes $O(N \log N)$ time per node, for a total preprocessing time of $O(N^2 \log N)$.
 During the algorithm, infeasible rings exceeding $(f/100) \cdot d^{\max}$ can then be discarded in $O(1)$ time by traversing this ordered structure.
 Since the preprocessing is performed only once, the overall complexity of \textsc{TSP-ep-all} remains unchanged.
\end{itemize}

\section{The Iterative Chainlet Partitioning Algorithm}
\label{sec:icp}

\begin{figure}
    \centering
    \includegraphics[width=\textwidth]{figures/ICP_Iteration.pdf}
    \caption{Illustrative Example of ICP Iteration}
    \label{fig:icp_iteration}
\end{figure}

This section presents the Iterative Chainlet Partitioning (ICP) algorithm. 
Figure \ref{fig:icp_iteration} illustrates the overall iteration process of the ICP algorithm. 
The algorithm begins by dividing the current chain into individual rings, which are then grouped into overlapping chainlets of manageable size.
Each chainlet is evaluated using the \textsc{TSP-ep-all} heuristic to determine its potential for improvement. 
The chainlet with the highest potential improvement is selected to update the corresponding segment of the chain, and the process iterates until no further improvements are possible.
Although the description focuses on the TSP-D, ICP readily applies to variants with customer eligibility and flying range limitations, as it inherits this capability from the \textsc{TSP-ep} and the \textsc{TSP-ep-all} subroutine.

The pseudo code of the ICP algorithm is presented in Appendix~\ref{sec:pseudo}. 
It begins by constructing an optimal TSP tour using the Concorde solver \citep{applegate1998solution}, which is then transformed into the initial chain using the \textsc{TSP-ep} heuristic. 
Rather than grouping a fixed number of rings into each chainlet, the algorithm dynamically adjusts the number of rings based on the chainlet's size, ensuring that each chainlet remains within a maximum size of 20. 
The \textsc{Group} process begins by sequentially adding rings to the current chainlet until adding another ring causes the chainlet to exceed the size limit. 
Once the chainlet reaches this limit, a new chainlet is initiated. 
If the newly generated chainlet is a subset of the previously generated chainlet, it is omitted from further consideration. 
The process then advances the starting ring to the next in sequence and continues until all rings have been grouped.
Adjusting the number of rings per chainlet ensures consistent algorithmic performance across various settings, as it controls the maximum chainlet size regardless of how problem parameters affect individual ring sizes.
The selection of this maximum node size balances several factors: grouping too many rings in a chainlet can slow the algorithm due to the high complexity of \textsc{TSP-ep-all}, while grouping too few may reduce the subroutine’s effectiveness.

We leverage a cache that stores previous \textsc{TSP-ep-all} results to avoid redundant calculations.
Since only the chainlet with the most significant improvement is updated in each iteration, the other non-overlapping chainlets remain unchanged in subsequent iterations.
Therefore, for previously encountered chainlets, the optimized chainlet and its corresponding improvement are retrieved directly from the cache, significantly enhancing the algorithm's efficiency.
If a chainlet is not in the cache, an input path is first constructed using a farthest insertion (FI) of the internal nodes.
The chainlet is then optimized via the \textsc{TSP-ep-all} subroutine, and the result is stored in the cache for future use.
Chainlets are hashed by node sequence, and while this hash is not perfectly unique, it is sufficient to guarantee consistent optimization.
The maximum node size and use of the Concorde solver and the FI heuristic are based on experimental results in Appendix \ref{sec:config}.

Now we formalize properties of the ICP algorithm. 
The following proposition highlights its deterministic structure, monotonic decrease in solution cost, and guaranteed termination in finite time. 
These properties are direct consequences of the algorithm's design: the greedy selection strategy ensures a strict cost reduction at each step, while the use of deterministic subroutines prevents the algorithm from revisiting prior configurations.

\begin{proposition} \label{prop:ICP_termination}
The ICP algorithm terminates in finite time, and the cost of the solution monotonically decreases throughout the iterations. 
In addition, if the instance has a unique TSP optimal solution, the ICP algorithm is deterministic; that is, every run of ICP for any given TSP-D instance results in the same solution.
\end{proposition}

The next proposition quantifies the efficiency of the ICP algorithm, derived from the structural properties in the formation and partitioning of rings and chainlets. 
Specifically, we first establish an upper bound on the number of rings possible in any chainlet and then combine this with the cache update rule to determine the maximum number of \textsc{TSP-ep-all} executions per iteration.

\begin{figure}
    \centering
        \includegraphics[width=1.0\textwidth]{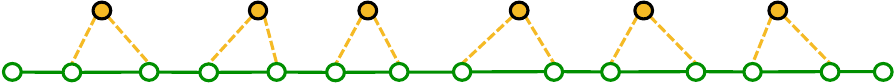}
    \caption{Chainlet with 20 nodes exhibiting the maximum number of rings}
    \label{fig:chainlet19}
\end{figure}

\begin{proposition} \label{prop:max_runs}
Suppose a chain constructed from a TSP-D instance with \(N\) total nodes where \(\alpha > 1\), no nodes are collinear, and \(N\) is sufficiently large. 
The maximum node size per chainlet in ICP is denoted as \(\ell\). 
Then, in the first iteration of ICP, the maximum number of \textsc{TSP-ep-all} runs is \(\lceil (2N-1)/3 \rceil\). 
In each subsequent iteration, the maximum number of \textsc{TSP-ep-all} runs is 
\begin{equation}
	\frac{4\ell - 9 + 2(\ell \bmod 3)}{3}.	\label{max_runs2}
\end{equation}
\end{proposition}

Proposition~\ref{prop:max_runs} indicates that in the initial iteration, the number of \textsc{TSP-ep-all} runs is linear in \(N\). 
However, this number remains constant in subsequent iterations, regardless of the instance size. 
With \(\ell=20\) in ICP, the necessary \textsc{TSP-ep-all} runs do not exceed 25 in later iterations. 

Our analysis in Section~\ref{subsec:ICP performance} substantiates and extends these theoretical bounds. 
We establish that \textsc{TSP-ep-all} constitutes the computationally costly component of ICP's runtime, particularly at higher \(\alpha\) values. 
Our empirical results further confirm that the number of iterations grows linearly with \(N\).
Note that since each \textsc{TSP-ep-all} execution operates on a small, size-constrained chainlet, its runtime is independent of the total problem size $N$.
Combining the linearity of iterations with our bound on subroutine calls per iteration, we conclude that the total number of \textsc{TSP-ep-all} executions scales as $O(N)$, ensuring ICP’s computational efficiency for large-scale instances.

\section{Neural Acceleration}
\label{sec:nicp}

This section details constructing a neural network model to predict the outcomes of the \textsc{TSP-ep-all} heuristic and its integration into the ICP algorithm to enhance computational efficiency.

Despite the efficiency of the ICP algorithm, a significant computational burden persists due to the need to compute \textsc{TSP-ep-all} for multiple chainlets in each iteration. 
To address this issue, we propose the integration of a neural cost predictor. 
This neural network serves as a surrogate for the cost, enabling the algorithm to bypass the direct execution of \textsc{TSP-ep-all} except for the chainlet with the highest predicted improvement. 
By leveraging GPU parallelization, the neural network can rapidly process multiple chainlets simultaneously, ensuring efficient prediction of potential improvements.
Consequently, this approach effectively reduces computational demands while preserving the integrity of the solution.

\subsection{Neural Cost Predictor} \label{subsec:NCP}

\begin{figure}
    \centering
    \includegraphics[width=\textwidth]{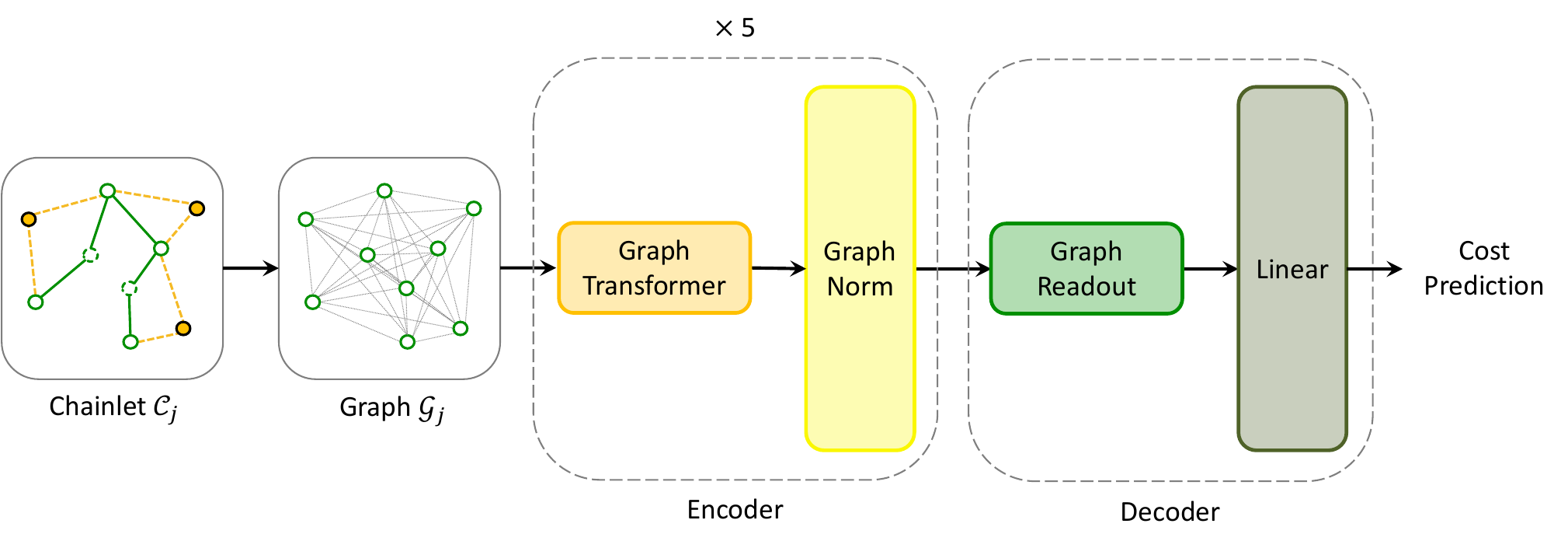}
    \caption{The neural cost predictor}
    \label{fig:nn}
\end{figure}

Figure \ref{fig:nn} depicts our neural network model. 
A GNN is chosen for its capability to effectively process input sizes that vary, such as the chainlets generated within the ICP algorithm.
Let graph \( \Gc = (\Vc, \Ec) \), where \( \Vc \) is the set of nodes with \( |\Vc| = n  \), and \( \Ec \) is the set of edges.
Each node \( v \in \Vc \) is associated with an initial feature vector \( \vec{X}_v \), and each edge \( e \in \Ec \) has a corresponding feature vector \( \vec{X}_e \). 
In a GNN, the representation of each node \( \vec{h}_v \) evolves through the network layers, starting with \( \vec{h}_v = \vec{X}_v \). 
The GNN iteratively updates these representations into \( \vec{h}_v' \) through message passing, wherein each node aggregates information from its neighbors, thereby capturing $k$-hop neighborhood interactions within the graph. 
The task is to learn the parameters \(\vec{\theta}\) of the cost prediction GNN \(g_\theta\), which estimates the \textsc{TSP-ep-all} cost \(\hat{y}\) for a given chainlet \(\mathcal{C}_j\) based on its graph representation \(\Gc_j\).
\[
\begin{aligned}
& \Gc_{j} \xleftarrow[\text{construction}]{\text{graph}} \mathcal{C}_{j} \\
& \hat{y} = g_\theta(\Gc_{j}) := \texttt{Decoder}(\texttt{Encoder}(\Gc_{j})),
\end{aligned}
\]

\paragraph*{Graph Construction}
The graph construction encapsulates the path sequence and truck and drone costs, precisely embedding the key inputs \textsc{TSP-ep-all} needs.
To represent a chainlet $\Cc_j$ as a graph $\Gc_j$, we construct a complete graph where nodes correspond to those in the chainlet. 
Node features are derived from a sinusoidal positional encoding of the initial path sequence generated by the FI heuristic. 
Edge features consist of the truck and drone travel costs, normalized by the maximum truck cost across all edges.

\paragraph*{Encoder}
The \texttt{Encoder} transforms the input graph into latent vectors using a stack of five identical layers. 
Each layer is composed of a graph transformer (GT) \citep{shi2020masked}, a graph-based adaptation of the Transformer \citep{vaswani2017attention} architecture; a specialized GraphNorm layer for normalization on graph data \citep{cai2021graphnorm}; and an ELU activation function \citep{clevert2015fast}. 
The transformation at each layer can be summarized as:
\begin{equation*}
\vec{h}_v' = \text{ELU}(\text{GraphNorm}(\text{GT}(\vec{h}_v))).
\end{equation*}

\paragraph*{Decoder}
The \texttt{Decoder} maps the latent node vectors $\vec{H'}$ to the final cost prediction $\hat{y}$. It first uses a graph readout module to produce a global graph vector by taking a softmax-weighted sum of the node features. This vector is then passed through a final linear layer to output the predicted cost:
\begin{equation*}
    \hat{y} = \mat{W}
    \bigg(\underbrace{\sum_{v \in \Vc} \text{softmax}\left(\vec{h}'_v \right) \vec{h}'_v}_{\text{graph readout}}\bigg) + b.
\end{equation*}

\subsection{Transfer Learning for Operational Constraints} \label{subsec:transfer_learning}

We now present a transfer learning strategy to efficiently extend the neural cost predictor to accommodate additional operational constraints.
 Our approach augments the frozen base model with a tailored FiLM layer in each block of the \texttt{Encoder}, which avoids costly retraining while preserving the original model's performance. 
We demonstrate this framework by extending the model to incorporate a finite drone flying range.

Note that in the graph transformer's edge processing, a zero value in a newly added edge feature dimension has no impact on the output, enabling a seamless identity with the original model. 
Leveraging this property, we introduce a normalized feature $f' = 1 - (f/200) \in [0,1]$. 
A value of $f'=0$ thus corresponds to the unlimited case.

The addition of this new edge feature alters the output of the preceding graph transformer layer, which in turn affects the feature statistics processed by the GraphNorm layer. 
A crafted FiLM layer, a generalization of conditional normalization methods \citep{perez2018film}, is therefore applied after the GraphNorm layer to adapt to these changes.
A FiLM layer performs a conditional, feature-wise affine transformation on the node embeddings:
\[ \text{FiLM}(\vec{h}_v \mid \vec{\gamma}, \vec{\beta}) = \vec{\gamma} \odot \vec{h}_v + \vec{\beta}, \]
where $\odot$ denotes the element-wise (Hadamard) product, and the per-channel vectors $\vec{\gamma}$ and $\vec{\beta}$ are generated from the feature $f'$. 
With a small multi-layer perceptron, $\phi$, we first compute a gating signal $\sigma = \phi(f') \cdot f'$. 
The parameters are then defined as:
\[ \vec{\gamma} = \vec{1} + \sigma \cdot \vec{s} \quad \text{and} \quad \vec{\beta} = \sigma \cdot \vec{t}, \]
where $\vec{s}$ and $\vec{t}$ are learnable per-channel vectors.
The gating signal $\sigma$ is designed to be zero when the flying range is unlimited ($f'=0$), resulting in $\vec{\gamma}=\vec{1}$ and $\vec{\beta}=\vec{0}$; the FiLM layer thus acts as an identity function. 
Consequently, extension is achieved by freezing the original model parameters and training only the lightweight new components: $\phi$, $\vec{s}$, $\vec{t}$, and the weights corresponding to the new edge feature dimension within each graph transformer. 

\subsection{The Neuro Iterative Chainlet Partitioning Algorithm}
\label{subsec:NICP}

We denote the ICP integrated with the neural cost predictor as Neuro Iterative Chainlet Partitioning (NICP).
NICP diverges from ICP by incorporating a neural network to predict the \textsc{TSP-ep-all} cost for each chainlet before applying the heuristic.
While substituting direct \textsc{TSP-ep-all} executions with neural predictions could diminish the benefit of caching, NICP maintains efficiency through a two-level caching mechanism that uses separate stores for optimized results and prediction data.

The pseudo code of the NICP algorithm is presented in Appendix~\ref{sec:pseudo}. 
NICP operates as follows: 
First, the algorithm checks if the results from a previous \textsc{TSP-ep-all} run are already available. 
If available, these results take priority, bypassing neural prediction. 
Otherwise, the algorithm searches for existing neural predictions. 
If found, it retrieves the cached input path and prediction. 
For chainlets without any cached information, the input path is constructed using an FI of the internal nodes.
The neural network predicts the \textsc{TSP-ep-all} cost using GPU parallelization for efficient processing of multiple chainlets, and the predicted cost is rescaled by multiplying the maximum truck cost across all edges to restore it to the original scale.
The expected improvement is subsequently derived from this rescaled cost, and cached along with the input path.

The chainlet with the highest improvement, actual or predicted, is selected for updating. If \textsc{TSP-ep-all} hasn't been applied to the chainlet, it is executed using the cached path, and both the improvement and chainlet are cached.
The chain is updated only if an improvement is confirmed.
The process continues until all chainlets have been processed by \textsc{TSP-ep-all} and no further improvements are possible, ensuring thorough exploration.

The key arguments in Proposition~\ref{prop:ICP_termination} for the ICP algorithm also apply to NICP. 
Although the neural predictor may yield imprecise estimates, every update is finalized only after \textsc{TSP-ep-all} confirms a strict improvement, ensuring finite termination. 
Moreover, for a given set of parameters \(\theta\), neural inference is deterministic, and hence NICP remains deterministic.

\section{Computational Study}
\label{sec:computational_study}

This section presents a computational study, including a performance evaluation of the ICP algorithm, neural network training, and a comparison between the ICP and NICP algorithms. 
All experiments were conducted on a workstation featuring an AMD Ryzen 9 5900X 12-Core Processor (3.7\,GHz, 24 logical processors), 64\,GB of DDR4 RAM at 3200\,MT/s, and an NVIDIA GeForce RTX~4070 GPU with 12\,GB of memory, running Ubuntu~22.04.3~LTS. 
The neural network was implemented and trained in Python 3.10, and the algorithms were developed in Julia 1.10.0, utilizing PyCall for cost predictions. 
The code is publicly available at \url{https://github.com/0505daniel/TSPDroneICP.jl}.

The evaluation utilized two benchmark sets. \emph{Set~A}, from \cite{agatz2018optimization}, includes 759 instances across three distribution types: uniform, 1-center (simulating a circular city center), and 2-center (creating two distinct clusters). 
These instances cover node sizes \(N\) ranging from 50 to 500 and truck-drone speed ratios \(\alpha \in \{1, 2, 3\}\). 
Additionally, to evaluate the TSP-D with flying range limitations, we include 190 instances with flying ranges $f \in [5, 50]$ percent, uniform distribution at a fixed $\alpha=2$, and $N \in \{50, 75, 100\}$. 
\emph{Set~B}, from \cite{bogyrbayeva2023deep}, consists of 300 instances at a fixed \(\alpha=2\), split between a random subset ($N \in \{50, 100\}$) and an Amsterdam-based subset ($N=50$) constructed from real-world locations to reflect realistic urban spatial patterns.  
Collectively, these benchmark sets enable comprehensive evaluation: Set~A provides systematic coverage of varying problem dimensions (node sizes, truck-drone speed ratios, and spatial distributions), while Set~B incorporates realistic urban spatial patterns. 
All instances contain at least 50 nodes, ensuring sufficient chainlet generation for thorough algorithmic evaluation.

We evaluated our algorithms against four benchmark algorithms: \textsc{TSP-ep-all} (dynamic programming), $\textsc{DPS}_{25}$ (divide-and-conquer), HM$_{4800}$ (end-to-end learning), and HGA-TAC$^{+}$ (metaheuristic).
Due to computational constraints, \textsc{TSP-ep-all} is applied only to Set~B. 
HM$_{4800}$ was trained explicitly for specific problem configurations, limiting its applicability; consequently, it is applied exclusively for Set~B. 
All benchmark algorithms were implemented in Julia and executed in the same computational environment.
For Set~A, we compare against $\textsc{DPS}_{25}$ and HGA-TAC$^+$, while for Set~B, we include all benchmark algorithms. 
The following subsections present summary results, with detailed outcomes provided in Appendix~\ref{sec:detailed_results}.

\subsection{Neural Network Training and Generalization}
\label{subsec:NN Training}

\begin{figure}
    \centering
    \begin{minipage}[t]{0.48\textwidth}
        \centering
        \includegraphics[width=\textwidth]{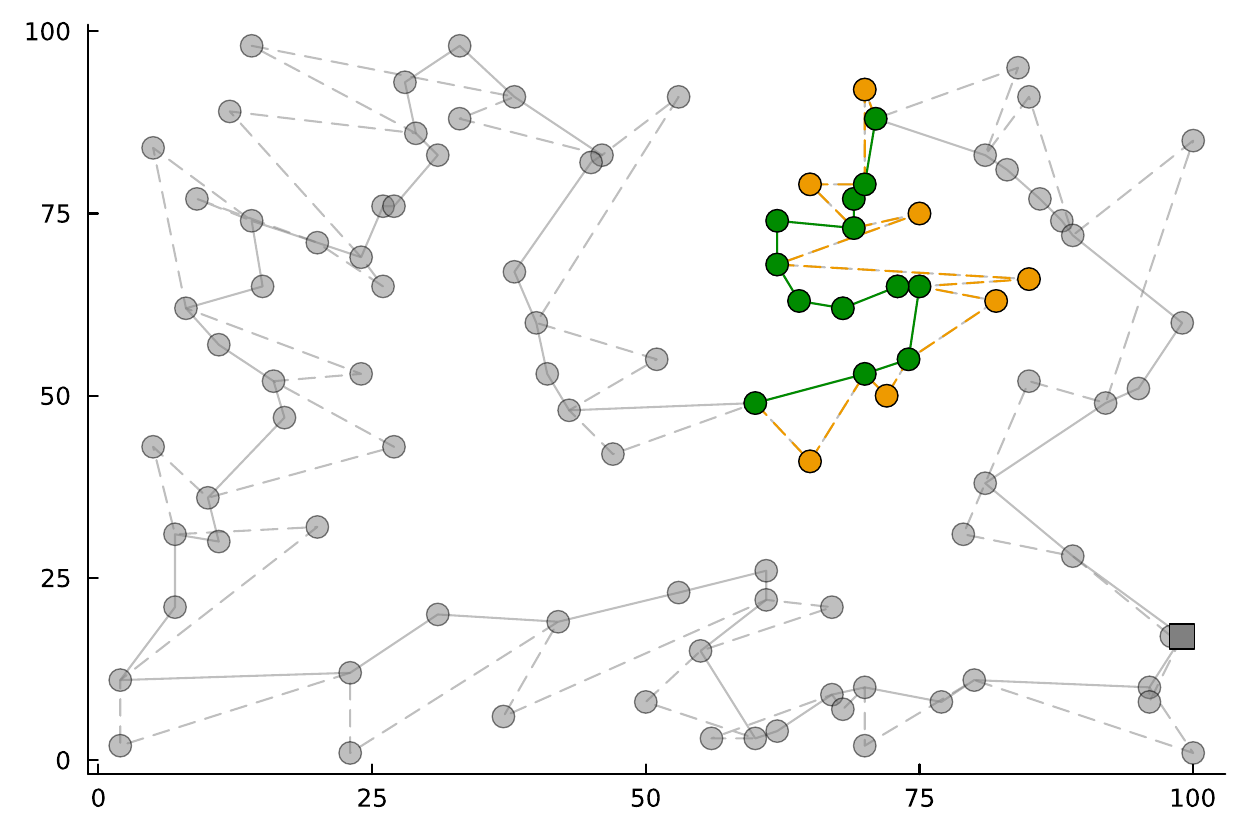}
        \caption{Distribution of nodes within chainlets from ICP algorithm runs.}
        \label{fig:chainlet_distribution}
    \end{minipage} %
    \hfill
    \begin{minipage}[t]{0.48\textwidth}
        \centering
                \includegraphics[width=\textwidth]{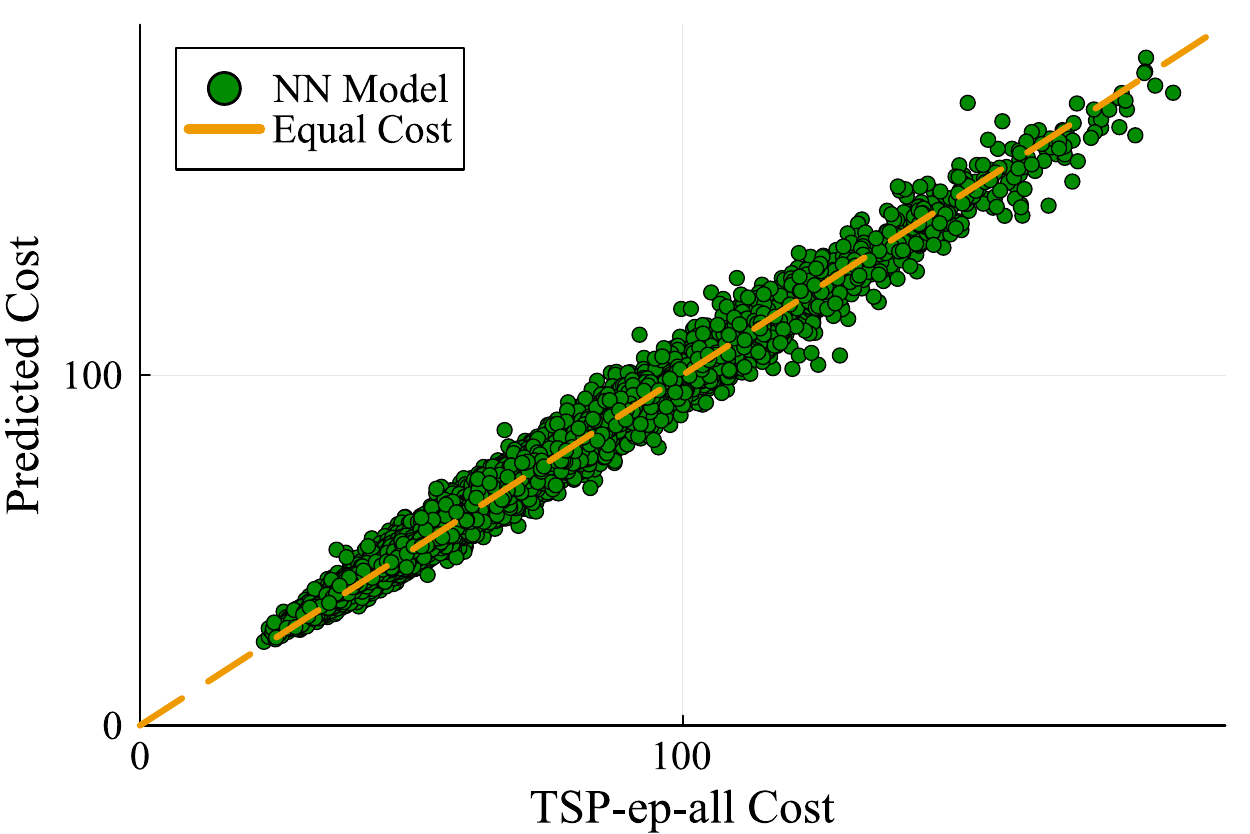}
        \caption{Predicted vs. actual costs for test set with uniform distribution.}
        \label{fig:scatter_uniform}
    \end{minipage}
\end{figure}

Creating training data that reflects the structure of chainlets in ICP is nontrivial, since chainlets are formed during optimization and often contain spatially clustered nodes. 
To capture these characteristics, we generated data by running ICP on uniformly distributed instances with $N \in [50, 500]$. 
This process yielded 384,000 samples (128,000 for each $\alpha \in {1,2,3}$). 
The model was trained for 100 epochs to predict the \textsc{TSP-ep-all} cost by minimizing mean squared error loss. 
Training required approximately 7 hours, confirming the practical feasibility of the learning phase.
We tested its performance on an independent test set of 38,400 data points.
To assess generalization, we also created OOD test sets for the 1-center, 2-center, and Amsterdam-based instances. 
For the Amsterdam set, coordinates were sampled from probability density functions derived from real-world locations using Gaussian kernel density estimation \citep {bogyrbayeva2023deep}.
Each test set contained 12,800 samples, equally divided across $\alpha \in \{1,2,3\}$.

\begin{table}
    \centering
    \caption{MAPE (\%) for test sets with different distributions.}
    \label{tab:nn_test}
\begin{tabular}{>{\centering\arraybackslash}p{1.0cm} |
                >{\centering\arraybackslash}p{1.5cm}
                >{\centering\arraybackslash}p{1.5cm}
                >{\centering\arraybackslash}p{1.5cm}
                >{\centering\arraybackslash}p{1.5cm} |
                >{\centering\arraybackslash}p{1.2cm}} %
        \toprule
        $\alpha$ & uniform & 1-center & 2-center & Ams. & Avg. \\
        \midrule
        1  & 3.33 & 4.64 & 4.96 & 4.24 & 4.29 \\
        2  & 2.95 & 3.34 & 3.65 & 3.74 & 3.42 \\
        3  & 3.75 & 5.01 & 5.40 & 4.80 & 4.74 \\
        \midrule
        Avg.  & 3.34 & 4.33 & 4.67 & 4.26 & 4.15 \\
        \bottomrule
    \end{tabular}
\end{table}

The neural cost predictor demonstrates high accuracy and robust generalization. 
Figure~\ref{fig:scatter_uniform} illustrates the close alignment of predicted and actual costs for uniform test instances.
Table~\ref{tab:nn_test} quantifies this, reporting a low Mean Absolute Percentage Error (MAPE) of 3.34\% on the uniform distribution. 
The model also generalizes effectively to unseen data, with only a minor increase in MAPE for the OOD 1-center (4.33\%), 2-center (4.67\%), and Amsterdam (4.26\%) distributions.
Moreover, performance remains consistent across relative speed ratios, with average MAPEs of 4.29\%, 3.42\%, and 4.74\% for $\alpha= 1, 2, 3$, respectively. 
In summary, constraining the predictor to small, fixed-size chainlets enables it to learn the essential local structural patterns. 
As a result, even when trained solely on uniform instances, it generalizes reliably across different spatial distributions and maintains consistent performance across varying values of $\alpha$, avoiding costly retraining.

To demonstrate the transfer learning framework described in Section~\ref{subsec:transfer_learning}, we trained the flying range extension using 128,000 samples with $f \in [5, 50]$, $N \in [50, 100]$, and $\alpha = 2$. Freezing the base model parameters, training took only 35 minutes due to its lightweight nature.
The extended model achieved an MAPE of 4.89\% on 12,800 test samples, demonstrating effective extension to the new constraint while preserving base model performance. 
Training progress curves and additional scatter plots for OOD distributions and transfer learning are provided in Appendix~\ref{sec:training_and_performance}.

\subsection{Computational Profile of ICP}
\label{subsec:icp_profile}

\begin{table}
\centering
\caption{Computational profile of ICP across main subroutines, averaged over 100 instances generated under Set~A's \citep{agatz2018optimization} uniform distribution scheme, for varying \(\alpha\) and \(N\). Times are in seconds.}
\label{tab:icp_profile}

\label{tab:icp_profile}
\begin{tabular}{
c  %
c  %
R{0.9cm} %
R{0.9cm} %
R{0.9cm} %
R{0.8cm} %
R{0.9cm}  %
R{0.9cm}  %
R{1.2cm} %
R{1.0cm} %
R{0.9cm} %
R{0.9cm} %
R{0.9cm} %
}
\toprule
& & 
\multicolumn{2}{c}{\textsc{Concorde}}
& \multicolumn{2}{c}{\textsc{TSP-ep}}
& \multicolumn{2}{c}{}
& \multicolumn{4}{c}{\textsc{TSP-ep-all}}
& \multicolumn{1}{c}{} \\
\cmidrule(lr){3-4}\cmidrule(lr){5-6}\cmidrule(lr){9-12}
\multicolumn{1}{c}{\(\alpha\)}
& \multicolumn{1}{c}{\(N\)}
& \multicolumn{1}{c}{Time}
& \multicolumn{1}{c}{\%}
& \multicolumn{1}{c}{Time}
& \multicolumn{1}{c}{\%}
& \multicolumn{1}{c}{\(|\Cc_j|\)}
& \multicolumn{1}{c}{\( \Ic \)}
& \multicolumn{1}{c}{\#}
& \multicolumn{1}{c}{Avg.}
& \multicolumn{1}{c}{Time}
& \multicolumn{1}{c}{\%}
& \multicolumn{1}{c}{\%} \\
\midrule
\multirow{5}*{1} 
& 100 & 0.14 & 41.01 & 0.000 & 0.04 & 18.05 & 5.99 & 29.70 & 0.007 & 0.20 & 57.32 & 98.38 \\
& 200 & 0.61 & 60.37 & 0.000 & 0.05 & 18.09 & 11.81 & 64.21 & 0.006 & 0.40 & 39.14 & 99.55 \\
& 300 & 1.95 & 75.55 & 0.001 & 0.06 & 18.09 & 18.13 & 101.47 & 0.006 & 0.62 & 24.03 & 99.63 \\
& 400 & 5.11 & 86.34 & 0.003 & 0.05 & 18.08 & 22.74 & 131.02 & 0.006 & 0.79 & 13.34 & 99.73 \\
& 500 & 9.57 & 90.17 & 0.005 & 0.05 & 18.11 & 28.69 & 167.17 & 0.006 & 1.01 & 9.53 & 99.75 \\
\midrule
\multirow{5}*{2}
& 100 & 0.09 & 7.31 & 0.000 & 0.01 & 19.36 & 8.77 & 94.41 & 0.012 & 1.15 & 92.16 & 99.49 \\
& 200 & 0.65 & 20.07 & 0.001 & 0.02 & 19.37 & 17.75 & 208.45 & 0.012 & 2.57 & 79.48 & 99.56 \\
& 300 & 2.40 & 38.51 & 0.001 & 0.02 & 19.37 & 25.98 & 312.84 & 0.012 & 3.80 & 60.94 & 99.47 \\
& 400 & 6.91 & 57.16 & 0.003 & 0.02 & 19.37 & 35.22 & 424.45 & 0.012 & 5.11 & 42.31 & 99.49 \\
& 500 & 8.69 & 57.31 & 0.005 & 0.03 & 19.37 & 43.79 & 535.97 & 0.012 & 6.38 & 42.04 & 99.38 \\
\midrule
\multirow{5}*{3}
& 100 & 0.16 & 6.91 & 0.000 & 0.01 & 19.44 & 10.09 & 125.11 & 0.017 & 2.10 & 92.87 & 99.79 \\
& 200 & 0.61 & 11.81 & 0.000 & 0.01 & 19.45 & 20.96 & 280.03 & 0.016 & 4.54 & 87.78 & 99.60 \\
& 300 & 1.78 & 20.28 & 0.001 & 0.01 & 19.45 & 31.28 & 431.67 & 0.016 & 6.95 & 79.15 & 99.44 \\
& 400 & 4.21 & 31.37 & 0.003 & 0.02 & 19.45 & 41.47 & 573.80 & 0.016 & 9.13 & 67.98 & 99.37 \\
& 500 & 9.65 & 46.13 & 0.005 & 0.03 & 19.45 & 52.62 & 723.24 & 0.015 & 11.14 & 53.24 & 99.39 \\
\bottomrule
\end{tabular}
\end{table}

\begin{figure}
   \centering
   \begin{subfigure}[b]{0.48\textwidth}
       \centering
       \includegraphics[width=\textwidth]{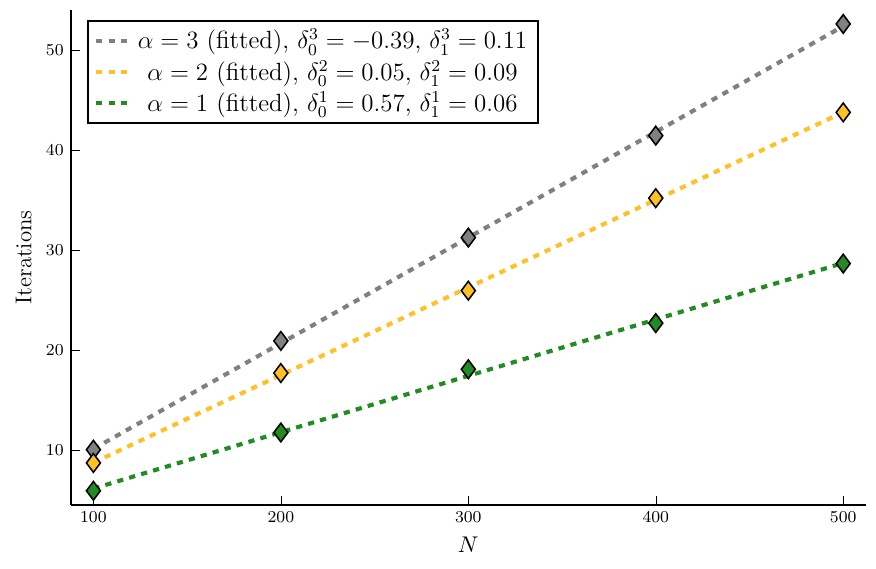}
       \caption{Number of iterations}
       \label{fig:iterations_vs_N}
   \end{subfigure}
   \hfill
   \begin{subfigure}[b]{0.48\textwidth}
       \centering
       \includegraphics[width=\textwidth]{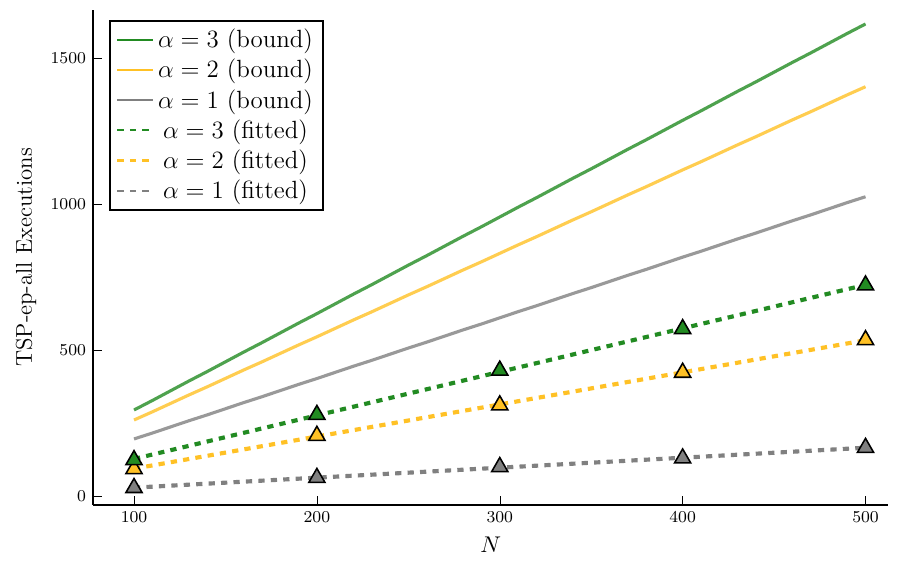}
       \caption{Number of \textsc{TSP-ep-all} executions with bounds}
       \label{fig:executions_with_bounds}
   \end{subfigure}   
      \caption{Linear growth patterns in ICP.}
   \label{fig:regression}
\end{figure}

We first examine ICP's computational characteristics to understand its scalability and the rationale for neural acceleration, with results detailed in Table~\ref{tab:icp_profile} and Figure~\ref{fig:regression}.
The table details the runtime of the algorithm's three principal subroutines: \textsc{Concorde}, \textsc{TSP-ep}, and \textsc{TSP-ep-all}. 
The data represent averages from 100 instances generated under Set~A's uniform distribution scheme. 
The columns detail the runtime components (Time) and the percentage contributions (\%) for each subroutine.
Additionally, for \textsc{TSP-ep-all}, we report the average chainlet size ($|\Cc_j|$), the number of iterations ($\Ic$), the total number of executions ($\#$), and the mean execution time per call (\textit{Avg.}).
The final column (\%) presents the total percentage contributions of these three subroutines, confirming that they account for nearly all of ICP's total runtime.

The results reveal several key patterns. 
A higher \(\alpha\) value results in larger average chainlet sizes and more frequent executions of \textsc{TSP-ep-all}, as faster drone speeds produce smaller rings and thus generate more chainlets per iteration.
Moreover, since \textsc{TSP-ep-all} is inherently faster for lower \(\alpha\), the average time per call and its share of total runtime rise significantly with \(\alpha\). 
This leads to a shift in the computational bottleneck: at low drone speeds, \textsc{Concorde} is the dominant component, while as the drone gets faster, the \textsc{TSP-ep-all} subroutine's runtime share grows significantly.

Figure~\ref{fig:iterations_vs_N} shows a strong linear relationship between the number of iterations and \(N\), confirmed by linear regression. 
This empirical result, when combined with Proposition~\ref{prop:max_runs}, confirms that the total number of \textsc{TSP-ep-all} executions is also bounded by a linear function of $N$. 
Figure~\ref{fig:executions_with_bounds} visualizes this theoretical bound (solid lines) alongside the actual number of executions (dashed lines), which also show clear linear growth. 
Consequently, as $N$ increases, the asymptotic complexity of ICP is dominated by the initial chain generation. 
Despite this favorable scaling, \textsc{TSP-ep-all} remains the costly component of total runtime for large instances, particularly at higher $\alpha$ values, highlighting the benefit of neural acceleration.

\subsection{Performance of ICP and Neural Acceleration}
\label{subsec:ICP performance}

This subsection presents ICP and NICP's performance from multiple perspectives, offering a comprehensive view of their behavior across the benchmark instances in Set~A and Set~B.
For Set~A, we used $\textsc{DPS}_{25}$ as the baseline algorithm, while for Set~B, we employed HM$_{4800}$ as the baseline algorithm.
We select HGA-TAC$^+$ as our primary benchmark due to its established record of high-quality solutions in the literature. 
For HGA-TAC$^+$, we report the Best, Mean, and Worst values from 10 independent runs.
Objective values are presented as relative gaps to the corresponding baseline algorithm.

\begin{table}
\centering
\caption{
    A summary of results for ICP and NICP across all instance sets. Gaps are relative to the baseline algorithm for each set.
    $^\dag$ is reported by \citet{bogyrbayeva2023deep}. Times are in seconds.
}
\label{tab:combined_results_distribution}
\newcolumntype{R}[1]{>{\raggedleft\arraybackslash}m{#1}}
\resizebox{\textwidth}{!}{%
\begin{tabular}{
    >{\raggedright\arraybackslash}m{2.2cm}  %
    R{0.9cm}                             %
    R{1.2cm}                             %
    R{1.2cm}                             %
    R{1.1cm}                             %
    R{0.8cm}                             %
    R{1.2cm}                             %
    R{0.8cm}                             %
    R{1.2cm}                             %
    R{0.8cm}                             %
    R{1.0cm}                             %
    R{1.4cm}                             %
}
    \toprule
    & \multicolumn{1}{c}{Baseline} & \multicolumn{4}{c}{HGA-TAC$^+$} & \multicolumn{2}{c}{ICP} & \multicolumn{2}{c}{NICP} & \multicolumn{2}{c}{Comparison (\%)} \\
    \cmidrule(lr){2-2}\cmidrule(lr){3-6}\cmidrule(lr){7-8}\cmidrule(lr){9-10}\cmidrule(lr){11-12}
    Instance Set & Time & Best & Mean & Worst & Time & Gap & Time & Gap & Time & Obj & Time \\
    \midrule
    \textbf{Set A} & \multicolumn{1}{c}{\textsc{DPS}$_{25}$} & & & & & & & & & & \\
    \cmidrule(lr){2-2}
    - Uniform & 2.56 & -1.02\% & 0.08\% & 1.27\% & 66.97 & -2.96\% & 5.03 & -2.84\% & 3.74 & 0.13\% & -25.81\% \\
    - 1-center & 1.82 & -1.38\% & -0.02\% & 1.55\% & 63.93 & -3.55\% & 4.27 & -3.45\% & 3.34 & 0.10\% & -21.74\% \\
    - 2-center & 1.80 & -1.12\% & 0.19\% & 1.58\% & 64.96 & -3.01\% & 4.14 & -2.84\% & 3.09 & 0.17\% & -25.37\% \\
    - Limited & 0.19 & 0.01\% & 0.57\% & 1.17\% & 9.61 & -1.21\% & 0.82 & -1.16\% & 0.62 & 0.05\% & -23.92\% \\
    \midrule
    \textbf{Set B} & \multicolumn{1}{c}{HM$_{4800}$} & & & & & & & & & & \\
    \cmidrule(lr){2-2}
    - Random & 8.96 & -1.30\% & 0.39\% & 2.17\% & 7.30 & -1.30\% & 1.02 & -1.14\% & 0.59 & 0.16\% & -42.10\% \\
    - Amsterdam & 1.41$^\dag$ & -1.79\% & 0.66\% & 2.97\% & 3.91 & -0.76\% & 0.47 & -0.44\% & 0.27 & 0.33\% & -43.06\% \\
    \bottomrule
\end{tabular}
}
\end{table}

\begin{table}
\centering
\caption{A summary of results for ICP and NICP with varying parameters $\alpha$ and $N$ for Set A unlimited instances. Times are in seconds.}
\label{tab:combined_results_alpha_N}
\resizebox{\textwidth}{!}{%
\begin{tabular}{
    >{\centering\arraybackslash}m{0.8cm}   %
    >{\centering\arraybackslash}m{0.5cm}   %
    R{0.9cm}                             %
    R{1.2cm}                             %
    R{1.2cm}                             %
    R{1.2cm}                             %
    R{0.9cm}                             %
    R{1.2cm}                             %
    R{0.8cm}                             %
    R{1.2cm}                             %
    R{0.8cm}                             %
    R{1.0cm}                             %
    R{1.4cm}                             %
}
    \toprule
    & & \multicolumn{1}{c}{\textsc{DPS}$_{25}$} & \multicolumn{4}{c}{HGA-TAC$^+$} & \multicolumn{2}{c}{ICP} & \multicolumn{2}{c}{NICP} & \multicolumn{2}{c}{Comparison (\%)} \\
    \cmidrule(lr){3-3}\cmidrule(lr){4-7}\cmidrule(lr){8-9}\cmidrule(lr){10-11}\cmidrule(lr){12-13}
    \textbf{Set A} & & Time & Best & Mean & Worst & Time & Gap & Time & Gap & Time & Obj & Time \\
    \midrule
    \multirow{3}{*}{$\alpha$}
    & 1 & 1.78 & -0.74\% & -0.33\% & 0.01\% & 46.43 & -1.27\% & 1.93 & -1.22\% & 1.98 & 0.06\% & 2.53\% \\
    & 2 & 2.10 & -0.27\% & 1.21\% & 2.87\% & 65.83 & -3.15\% & 4.56 & -3.04\% & 3.29 & 0.12\% & -27.78\% \\
    & 3 & 2.31 & -2.78\% & -0.55\% & 1.97\% & 83.69 & -5.71\% & 6.98 & -5.45\% & 4.91 & 0.27\% & -29.64\% \\
    \midrule
    \multirow{7}{*}{$N$}
    & 50 & 0.16 & -3.87\% & -2.20\% & -0.29\% & 4.37 & -2.41\% & 0.49 & -2.30\% & 0.33 & 0.11\% & -33.08\% \\
    & 75 & 0.27 & -3.24\% & -1.45\% & 0.44\% & 8.14 & -2.56\% & 0.92 & -2.52\% & 0.58 & 0.04\% & -36.63\% \\
    & 100 & 0.34 & -2.37\% & -0.68\% & 0.97\% & 12.83 & -3.01\% & 1.31 & -2.80\% & 0.86 & 0.22\% & -34.06\% \\
    & 175 & 1.14 & -1.47\% & -0.07\% & 1.33\% & 31.43 & -2.91\% & 2.90 & -2.78\% & 2.04 & 0.13\% & -29.68\% \\
    & 250 & 1.41 & -1.08\% & 0.19\% & 1.40\% & 58.52 & -3.36\% & 4.07 & -3.24\% & 2.87 & 0.12\% & -29.57\% \\
    & 375 & 3.74 & -0.47\% & 0.56\% & 1.85\% & 120.28 & -3.36\% & 7.86 & -3.20\% & 5.86 & 0.17\% & -25.50\% \\
    & 500 & 7.22 & 0.03\% & 1.11\% & 2.47\% & 202.59 & -3.39\% & 12.69 & -3.27\% & 10.56 & 0.13\% & -16.83\% \\
    \bottomrule
\end{tabular}
}
\end{table}

Overall, ICP establishes a new state-of-the-art, achieving an average solution quality improvement of 2.6\% compared to the mean performance of HGA-TAC$^+$ while reducing computation time by an average of 91.3\%.
Moreover, NICP successfully improves efficiency, reducing computation time by 28.6\% with a negligible objective value increase of only 0.14\% compared to ICP. 

Table~\ref{tab:combined_results_distribution} summarizes the performance across different distributions.
Across both Set~A and Set~B, ICP consistently outperforms the state-of-the-art HGA-TAC$^+$, improving upon its mean results while being significantly faster.
Notably, for Set~A instances, ICP often surpasses even the \emph{best} solution obtained from 10~runs.
This strong performance extends to more realistic variants with flying range limitations.
The table further shows that neural acceleration successfully preserves these gains.
NICP maintains ICP's high solution quality with minimal objective degradation (0.05\% to 0.33\%) while achieving substantial time savings (21.7\% to 43.1\%).
Its robust performance across various spatial distributions supports the neural predictor's generalization to out-of-distribution instances, while its performance on limited-range instances validates our transfer learning framework.

Table~\ref{tab:combined_results_alpha_N} examines scalability across varying speed ratios ($\alpha$) and instance sizes ($N$). 
The results show that ICP's advantage in solution quality over HGA-TAC$^+$ grows with both parameters. 
In contrast to the steep runtime growth of HGA-TAC$^+$, ICP exhibits more moderate scaling, increasing its time savings on larger instances. 
Turning to the neural acceleration, NICP maintains consistent, high-quality solutions across all $\alpha$ and $N$ values. 
Its impact on runtime is most pronounced at higher drone speeds, aligning with our profiling analysis; while slightly slower at $\alpha=1$, it yields substantial time reductions for $\alpha=2$ and $\alpha=3$, where the computational burden of \textsc{TSP-ep-all} is higher.
Regarding instance size, the time savings decrease as the relative contribution of \textsc{TSP-ep-all} to total runtime decreases.
Nevertheless, it demonstrates a significant computational advantage across various problem settings, particularly at higher $\alpha$ values.

\begin{figure}
    \centering
    \begin{subfigure}[b]{0.48\textwidth}
        \centering
        \includegraphics[width=\textwidth]{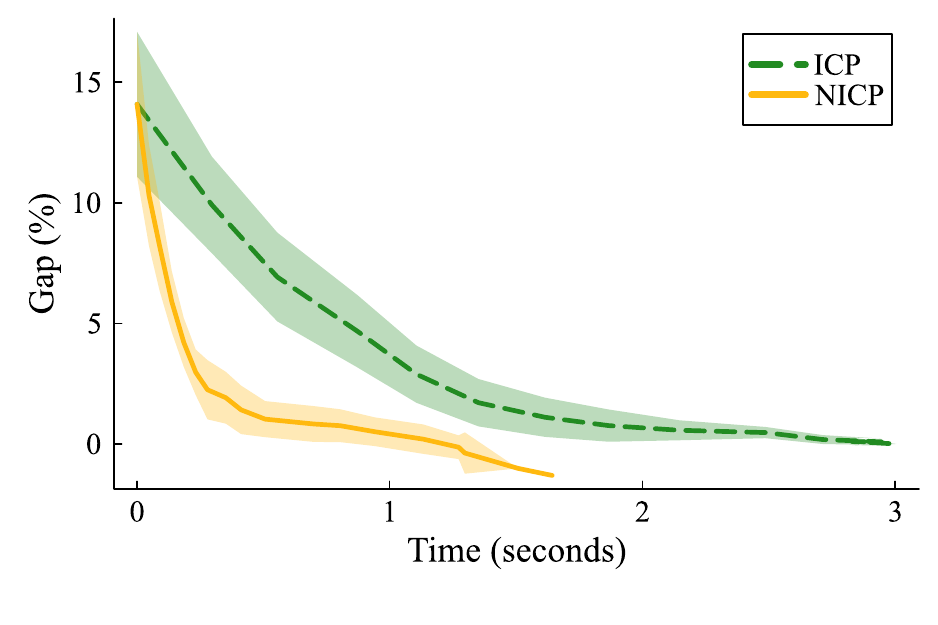}
        \caption{Objective Value over Time}
        \label{fig:time_vs_obj}
    \end{subfigure}
    \hfill
    \begin{subfigure}[b]{0.48\textwidth}
        \centering
                \includegraphics[width=\textwidth]{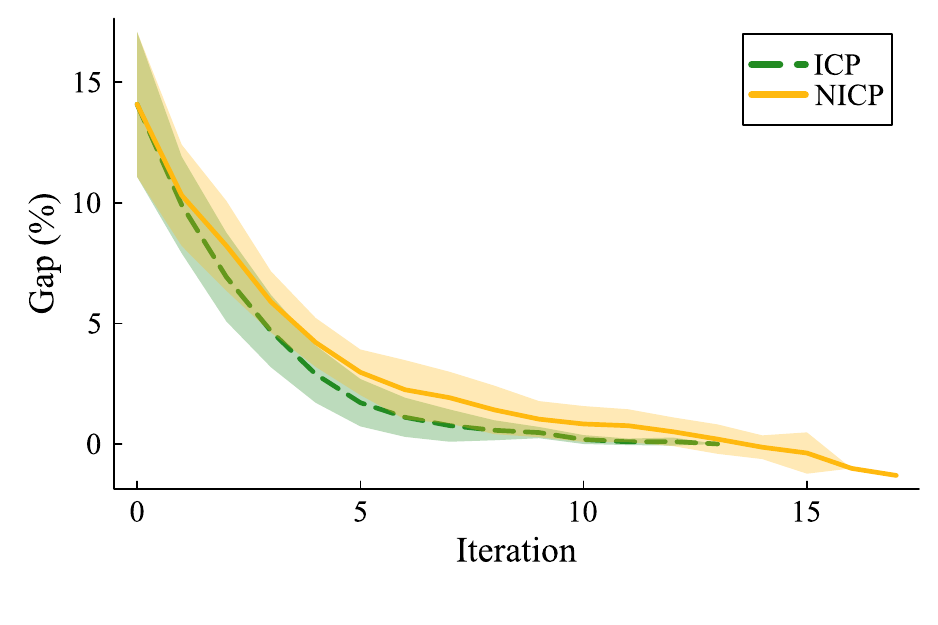}
        \caption{Objective Value over Iteration}
        \label{fig:iteration_vs_obj}
    \end{subfigure}
    \caption{Comparison of average ICP and NICP results for \( 10 \) randomly generated uniform instances with \( N = 100 \).}
    \label{fig:ICP_NICP_comparison}
\end{figure}

We conclude by illustrating the characteristics of neural acceleration through Figure~\ref{fig:ICP_NICP_comparison}, which compares ICP and NICP across ten uniformly generated instances with $N=100$ and $\alpha=2$.
The left figure presents the gap between ICP's final objective value over computation time, where the shaded regions represent standard deviations. 
NICP exhibits faster convergence, achieving greater reductions within the same time frame while closely tracking ICP's solution quality. 
Complementing this time-based analysis, the right figure shows the gap over iterations, demonstrating that NICP follows ICP's solution trajectory per iteration despite its faster execution. 
Together, these results validate that NICP achieves actual neural acceleration - maintaining ICP's solution patterns while significantly reducing computation time through efficient neural prediction.

\section{Conclusion}
\label{sec:conclusion}

In this work, we introduced the Iterative Chainlet Partitioning (ICP) algorithm, a deterministic and theoretically-grounded solution methodology for the TSP-D, along with its neural acceleration (NICP).
Our results demonstrate that ICP establishes a new state-of-the-art, significantly outperforming existing methods in both solution quality and computational speed across a wide range of benchmarks, including those with practical flying range limitations.
The deterministic nature of ICP ensures reliability, while its novel structure provides a rigorous theoretical basis for its scalability.

Furthermore, we showed that a targeted hybrid approach can provide substantial acceleration without compromising performance. 
By integrating a GNN predictor trained with a novel transfer learning strategy, NICP reduces computation time significantly while preserving ICP's high-quality solutions.
The adaptability of this framework makes it a valuable and extensible foundation for developing high-performance algorithms for a wide range of truck-drone routing problems.

Future work will focus on improving methods for initial chain generation, which currently dominates computational complexity for larger instances.
Additionally, applying the transfer learning framework to other operational constraints, such as customer eligibility restrictions, offers promising avenues for expansion.

\nobibliography*

\bibliographystyle{apalike}
\bibliography{library}

\newpage
\renewcommand{\appendixpagename}{Appendices}
\appendix
\appendixpage

\section{Proofs of Propositions and Lemmas} \label{sec:proofs}

\proof{Proof of Lemma \ref{lem:EP_long_ring_monotone}.}
\(E(i, k)\) is the first index with \(C^d(i, j, k) \leq C^t(i, j, k)\).
We prove that such property holds for any \(j\) greater than \(E(i, k)\) by mathematical induction.
Suppose \(C^d(i, j, k) \leq C^t(i, j, k)\). Then by the triangular inequality,
\begin{align*}
    C^d(i, j+1, k) &= c^d(p_i, p_k) + c^d(p_k, p_{j+1})\\
    &\leq c^d(p_i, p_k) + c^d(p_k, p_j) + c^d(p_j, p_{j+1})\\
    &= C^d(i, j, k) + c^d(p_j, p_{j+1})\\
    &\leq C^t(i ,j, k) + (1/\alpha) \cdot c^t(p_j, p_{j+1})\\
    &\leq C^t(i, j, k) + c^t(p_j, p_{j+1})\\
    &= \left[S^t(i, k-1) + S^t(k+1, j)+c^t(p_{k-1}, p_{k+1})\right] + c^t(p_j, p_{j+1})\\
    &= S^t(i, k-1) + S^t(k+1, j+1)+c^t(p_{k-1}, p_{k+1})\\
    &= C^t(i, j+1, k).
\end{align*}
which completes the proof. 
\endproof

\proof{Proof of Lemma \ref{lem:EP_num_short_ring}.}
By definition,
\[|\breve{R}^s| = \Big|\big\{(i, j, k) \in \breve{R} : j \leq E(i, k) \big\}\Big| = \sum_{i<k} \big[E(i, k)-k\big].\]
Note that as $\alpha$ grows, \(E(i, k)\) decreases and so does \(|\breve{R}^s|\).
Therefore, let us prove the theorem for the case of $\alpha=1$.
\begin{align*}
    |\breve{R}^s| &= \Big| \big\{(i, j, k) \in \breve{R} \mid j < E(i, k) \big\} \Big| + \Big| \big\{(i, j, k) \in \breve{R} \mid j = E(i, k)\big\} \Big|\\
    &< \Big| \big\{(i, j, k) \in \breve{R} \mid j < E(i, k) \big\} \Big| + \frac{1}{2} n^2\\
    &= \Big| \big\{(i, j, k) \in R \mid C^d(i, j, k) > C^t(i, j, k) \big\} \Big| + \frac{1}{2} n^2.
\end{align*}
Let a start node \(i\) and end node \(j\) be given.
We are going to prove that there is at most one \(k\) such that \(C^d(i, j, k) > C^t(i, j, k)\).
Suppose \(C^d(i, j, k_1) > C^t(i, j, k_1)\).
For any \(k_1 < k_2 < j\), the following holds by the triangular inequality.
\begin{align*}
C^d(i, j, k_2)
&= c^d(p_i, p_{k_2}) + c^d(p_{k_2}, p_j) = c^t(p_i, p_{k_2}) + c^t(p_{k_2}, p_j)\\
&\leq \left[S^t(i, k_1-1)+S^t(k_1+1, k_2) + c^t(p_{k_1-1}, p_{k_1+1})\right] + S^t(k_2, j)\\
&= S^t(i, k_1-1)+S^t(k_1+1, j)+c^t(p_{k_1-1}, p_{k_1+1})\\
&= C^t(i, j, k_1)\\
&< C^d(i, j, k_1)\\
&= c^d(p_i, p_{k_1}) + c^d(p_{k_1}, p_j) = c^t(p_i, p_{k_1}) + c^t(p_{k_1}, p_j)\\
&\leq S^t(i, k_1) + \left[S^t(k_1, k_2-1)+S^t(k_2+1, j)+c^d(k_2-1, k_2+1)\right]  \\
&= S^t(i, k_2-1)+S^t(k_2+1, j)+c^d(k_2-1, k_2+1)\\
&= C^t(i, j, k_2)
\end{align*}
Therefore for any node pair \(i<j\), there is at most one \(k\) such that \(C^d(i, j, k) > C^t(i, j, k)\).
\[
\Big| \big\{(i, j, k) \in R \mid C^d(i, j, k) > C^t(i, j, k) \big\} \Big| < \frac{1}{2} n^2.
\]
We conclude $|\breve{R}^s| < \frac{1}{2} n^2 + \frac{1}{2} n^2 = n^2$ for any $\alpha$, hence a proof.
\endproof

\proof{Proof of Proposition \ref{prop:ICP_termination}.}
The ICP algorithm ensures that in each iteration, the chainlet with the highest positive improvement is selected and updated.
This guarantees that the cost of the solution monotonically decreases over iterations, as no update can increase the cost.
Moreover, cycling cannot occur in ICP.
Once a chainlet is updated, each modified ring within the chainlet is an exact partitioning solution for the current path sequence.
For any chainlet containing these modified rings in subsequent iterations, the result of applying \textsc{TSP-ep-all} will remain unchanged unless the path sequence itself is modified.
Thus, rings modified during previous updates cannot appear in their original configuration in subsequent iterations.
As a result, the algorithm progresses without revisiting prior configurations.
Since the number of possible configurations of chainlets is finite, and each iteration guarantees a strict cost improvement, the ICP algorithm must terminate in a finite number of iterations.

Additionally, ICP is a deterministic procedure.
This follows from the deterministic nature of subroutines: the \textsc{Concorde} solver is deterministic given that the TSP instance has a unique optimal solution, and both \textsc{TSP-ep} and \textsc{TSP-ep-all} are deterministic.
Therefore, with the consistent greedy selection of chainlets, the sequence of chainlet updates depends only on the input instance and the deterministic subroutines, ensuring the final solution is uniquely determined.
\endproof

\begin{figure}[h]
    \centering
    \vspace{0.5cm}
    \includegraphics[width=0.3\textwidth]{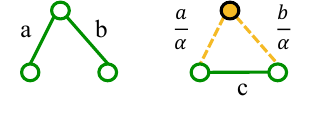}
    \caption{Two Consecutive \textit{Straight Rings} and an alternative \textit{Triangular Ring}}
    \label{fig:triangle}
\end{figure}

\begin{lemma} \label{lem:consecutive_straight_rings}
Suppose a TSP-D instance where $\alpha > 1$ and no nodes are collinear. Then, no two consecutive straight rings can appear in an optimal TSP-D solution. The same holds for exact partitioning solutions for any given path.
\end{lemma}

\proof{Proof of Lemma \ref{lem:consecutive_straight_rings}.}
Consider two consecutive straight rings with arc costs \(a\) and \(b\) connecting three consecutive combined nodes, illustrated in Figure~\ref{fig:triangle}. The corresponding triangular ring can be formed where a drone serves the middle node, and the truck directly connects the first and third combined nodes. Let \(c\) denote the cost of the arc directly connecting these two combined nodes in the triangular ring.
The total cost of the two consecutive straight rings is \(a + b\), while the total cost of the triangular ring is \(\max\left\{\frac{a + b}{\alpha}, c\right\}\).
If \(\frac{a + b}{\alpha} \geq c\), the cost of the triangular ring is \(\frac{a + b}{\alpha}\), and since \(\alpha > 1\), it follows that \(\frac{a + b}{\alpha} < a + b\).
If \(\frac{a + b}{\alpha} < c\), the cost of the triangular ring is \(c\), and by the triangle inequality, \(c < a + b\).
Thus, the cost of the triangular ring is strictly less than the cost of the two consecutive straight rings when \(\alpha > 1\).
\endproof

Lemma \ref{lem:consecutive_straight_rings} shows that for $\alpha > 1$ and non-collinear nodes, two consecutive straight rings do not appear in an optimal solution, as a single triangular ring provides a lower-cost alternative.
The result also holds for exact partitioning solutions, as exact partitioning guarantees optimality within the current TSP sequence.

The structural property given by Lemma~\ref{lem:consecutive_straight_rings} enables the calculation of the maximum number of rings in a chain.
Each ring must contain minimal nodes, achieved by maximizing the alternation between straight and triangular rings as shown in Figure \ref{fig:chainlet19}.
The following lemma formalizes this idea.

\begin{lemma}  \label{lem:max_rings}
Suppose a chain is constructed from a TSP-D instance where \(\alpha > 1\) and no nodes are collinear. Then, for any chainlet with \(n \geq 3\) nodes, the maximum number of rings is
\begin{equation}
	\frac{2n - 3 + (n \bmod 3)}{3} .
\end{equation}
\end{lemma}

\proof{Proof of Lemma \ref{lem:max_rings}.}
The maximum number of rings in a chainlet occurs when there is a minimal number of nodes in each ring, which is achieved by maximizing the alternation between a straight ring and a triangular ring.
Three cases are considered:

\begin{enumerate}[label=(\roman*)]
\item If \(n \equiv 1 \pmod{3}\), the maximum number of rings results from alternating between straight and triangular rings.
Starting from a start node, a straight ring followed by a triangular ring yields 2 rings for every 3 nodes, so the total number of rings is \( \frac{2(n-1)}{3} \).
 An equivalent number of rings is obtained by initiating the sequence with a triangular ring.

\item For a chainlet with one additional node than case (i), \textit{i.e.}, \(n \equiv 2 \pmod{3}\), adding a straight ring maximizes the number of rings, giving \( \frac{2(n-2)}{3} + 1 \).

\item For a chainlet with one fewer node than case (i), \textit{i.e.}, \(n \equiv 0 \pmod{3}\), attempting to change a triangular ring into a straight ring would always result in two consecutive straight rings.
Thus, the only choice is to delete a straight ring, yielding \( \frac{2n}{3} - 1 \).
\end{enumerate}
Altogether, the maximum number of rings in a chainlet can be universally stated as \( \frac{2n - 3 + (n \bmod 3)}{3} \).
\endproof

Lemma~\ref{lem:max_rings} provides the maximum number of rings in any chainlet.
Together with the caching mechanism—which only treats chainlets containing at least one ring from a modified chainlet as new—this result bounds the number of \textsc{TSP-ep-all} subroutine runs necessary per ICP iteration.

\proof{Proof of Proposition \ref{prop:max_runs}.}
The maximum number of rings in a chain sets an upper bound on the number of chainlets generated, as each chainlet created by \textsc{Group} must contain at least one ring that has not been grouped into any chainlet.
Unlike the count established in Lemma~\ref{lem:max_rings} for chainlets, a chain includes a dummy node representing the depot, which increases the overall node count by one.
Therefore, plugging in $N+1$ into the formula from Lemma~\ref{lem:max_rings}, we obtain \[ \frac{2(N + 1) - 3 + ((N + 1) \bmod 3)}{3} \] or \(\lceil (2N-1)/3 \rceil\) as the maximum number of rings in the chain.
This number also bounds the maximum number of chainlets in the first iteration.
It is also the maximum number of \textsc{TSP-ep-all} runs required in the first iteration where every chainlet is initially new.

In subsequent iterations, \textsc{TSP-ep-all} is executed only on chainlets containing rings modified by the updated chainlet.
Consider an updated chainlet $\mathcal{C}_k$ with $r$ rings.
Define the prefix set as the collection of all consecutive subsequences starting from the first ring, and the suffix set as the collection of all consecutive subsequences ending with the last ring.
Each set contains exactly $r$ sequences, with the full sequence appearing in both sets.
The union of the chainlet's prefix and suffix sets contains at most $2r - 1$ unique sequences.
The maximum number of new chainlets arises when $\Cc_k$ contains the maximum number of rings and each element in this union forms a distinct chainlet.

By Lemma~\ref{lem:max_rings}, the maximum number of rings in the updated chainlet is \( \frac{2l - (3 -  l \bmod 3)}{3} \).
Consequently, the maximum number of new chainlets is \( 2\left( \frac{2l - (3 - l \bmod 3)}{3} \right) - 1 \) or \eqref{max_runs2}, which is also the maximum number of \textsc{TSP-ep-all} runs required in subsequent iterations.
\endproof

\newpage

\section{Pseudo Codes} \label{sec:pseudo}

\renewcommand{\thealgorithm}{\arabic{algorithm}}
\setcounter{algorithm}{0} 

This appendix provides the detailed pseudocode for the ICP and NICP algorithms. Table~\ref{tab:pseudo_notation} defines the key notations used in the algorithms.

\begin{table}[h]
\scriptsize
\centering
\caption{Description of notations used in Algorithms 1 and 2.}
\label{tab:pseudo_notation}
\begin{tabular}{ll}
\toprule
\textbf{Notation} & \textbf{Description} \\
\midrule
$\mathcal{N}$ & The set of all customer nodes, including the depot. \\
$\Pc$ & A sequence of nodes $(p_1, \ldots, p_{N+1})$ representing the initial TSP tour. \\
$\mathcal{C}$ & A chain, representing a feasible TSP-D solution. \\
$\Pc_j$ & The input path constructed for the nodes within chainlet $\mathcal{C}_j$. \\
$\mathcal{C}_j$ & The $j$-th chainlet, a segment of the chain $\mathcal{C}$. \\
$\mathcal{C}_{j}^{\prime}$ & The optimized version of chainlet $\mathcal{C}_j$. \\
$\Delta_j$ & The actual cost improvement obtained from optimizing $\mathcal{C}_j$. \\
$\hat{\Delta}_j$ & The predicted cost improvement for $\mathcal{C}_j$ from the GNN. \\
\texttt{cache} & A cache storing results for chainlets in ICP. \\
\texttt{cache.optimized} & In NICP, a cache for results from \textsc{TSP-ep-all}. \\
\texttt{cache.predicted} & In NICP, a cache for GNN predictions and input paths. \\
\bottomrule
\end{tabular}
\end{table}

\begin{algorithm}[h]
\caption{Iterative Chainlet Partitioning (ICP)}
\label{alg:ICP}
\begin{algorithmic}[1]
\State \textbf{Input:} Customer Node Set $\mathcal{N}$
\State \textbf{Output:} Optimized Chain $\mathcal{C}$
\State $\Pc \gets \textsc{Concorde}(\mathcal{N})$
\State $\mathcal{C} \gets \textsc{TSP-ep}(\Pc)$
\State Initialize \texttt{cache} for chainlets
\Repeat
    \State $\mathcal{C}_1, \mathcal{C}_2, \ldots, \mathcal{C}_m \gets \textsc{Group}(\mathcal{C})$ \Comment{Group rings into chainlets}
    \For{$ j \in \{1, 2, \ldots, m\}$}
    \If{$\textsc{Hash}(\Cc_j) \in \texttt{cache}$ }
        \State $\mathcal{C}_{j}^{\prime}, \Delta_j \gets \texttt{cache}[\textsc{Hash}(\Cc_j)]$  
    \Else
        \State  $\Pc_j \gets \textsc{FarthestInsertion}(\mathcal{C}_j)$  \Comment{Farthest insertion with fixed end}  
        \State $\mathcal{C}_{j}^{\prime} \gets \textsc{TSP-ep-all}(\Pc_j)$  
        \State $\Delta_j \gets \texttt{Cost}(\mathcal{C}_j) - \texttt{Cost}(\mathcal{C}_{j}^{\prime})$ 
        \State $\texttt{cache}[\textsc{Hash}(\Cc_j)] \gets (\mathcal{C}_{j}^{\prime}, \Delta_j )$
        \State $\texttt{cache}[\textsc{Hash}(\Cc_j^\prime)] \gets (\mathcal{C}_{j}^{\prime}, 0)$ 
    \EndIf 
    \EndFor
    \State $k \gets \argmax_{i\in\{1,2,...,m\}} \Delta_i$
    \If{$\Delta_k > 0$}
        \State Replace $\mathcal{C}_k$ in $\mathcal{C}$ with $\mathcal{C}_k^{\prime}$
    \EndIf
\Until{$\Delta_i \leq 0 \ \forall i \in\{1, \ldots, m\}$}
\State \textbf{return} $\mathcal{C}$
\end{algorithmic}
\end{algorithm}

\begin{algorithm}[h]
\caption{Neuro Iterative Chainlet Partitioning (NICP)}
\label{alg:NICP}
\begin{algorithmic}[1]
\State \textbf{Input:} Customer Node Set $\mathcal{N}$
\State \textbf{Output:} Optimized Chain $\mathcal{C}$
\State $\Pc \gets \textsc{Concorde}(\mathcal{N})$
\State $\mathcal{C} \gets \textsc{TSP-ep}(\Pc)$
\State Initialize \texttt{cache.optimized} and \texttt{cache.predicted}
\Repeat
    \State $\mathcal{C}_1, \mathcal{C}_2, \ldots, \mathcal{C}_m \gets \textsc{Group}(\mathcal{C})$ \Comment{Group rings into chainlets}
    \For{$j \in \{1, 2, \ldots, m\}$}
        \If{$\textsc{Hash}(\mathcal{C}_j) \in \texttt{cache.optimized}$}
            \State $\mathcal{C}_{j}^{\prime}, \Delta_j \gets \texttt{cache.optimized}[\textsc{Hash}(\mathcal{C}_j)]$
            \State $\hat{\Delta}_j \gets \Delta_j$
        \ElsIf{$\textsc{Hash}(\mathcal{C}_j) \in \texttt{cache.predicted}$}
            \State $\Pc_j, \hat{\Delta}_j \gets \texttt{cache.predicted}[\textsc{Hash}(\mathcal{C}_j)]$
        \Else
            \State  $\Pc_j \gets \textsc{FarthestInsertion}(\mathcal{C}_j)$  \Comment{Farthest insertion with fixed end} 
            \State $\hat{\Delta}_j \gets \texttt{Predictor}(\Pc_j)$
            \State $\texttt{cache.predicted}[\textsc{Hash}(\mathcal{C}_j)] \gets (\Pc_j, \hat{\Delta}_j)$
        \EndIf
    \EndFor
    
    \State $k \gets \argmax_{i\in\{1,2,...,m\}} \hat{\Delta}_i$
    \If{$\textsc{Hash}(\mathcal{C}_k) \notin \texttt{cache.optimized}$}
        \State $\mathcal{C}_{k}^{\prime} \gets \textsc{TSP-ep-all}(\Pc_k)$
        \State $\Delta_k \gets \texttt{Cost}(\mathcal{C}_k) - \texttt{Cost}(\mathcal{C}_{k}^{\prime})$
        \State $\texttt{cache.optimized}[\textsc{Hash}(\mathcal{C}_k)] \gets (\mathcal{C}_k^{\prime}, \Delta_k)$
        \State $\texttt{cache.optimized}[\textsc{Hash}(\mathcal{C}_k^\prime)] \gets (\mathcal{C}_k^{\prime}, 0)$
    \EndIf
    
    \If{$\Delta_k > 0$}
        \State Replace $\mathcal{C}_k$ in $\mathcal{C}$ with $\mathcal{C}_k^{\prime}$
    \EndIf
\Until{$\forall i \in\{1, \ldots, m\}: \Delta_i \leq 0 \text{ and } \textsc{Hash}(\mathcal{C}_i) \in \texttt{cache.optimized}$}
\State \textbf{return} $\mathcal{C}$  
\end{algorithmic}
\end{algorithm}

\newpage
\section{Selection of ICP Configurations} \label{sec:config}

\begin{figure}[h]
   \centering
   \begin{minipage}{0.48\textwidth}
       \centering
       \includegraphics[width=\textwidth]{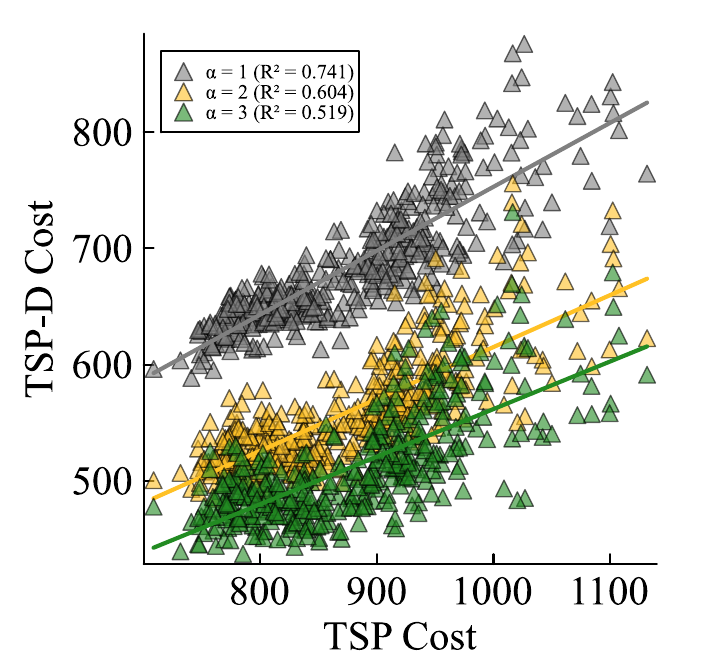}
       \caption{Linear regression analysis of TSP versus TSP-D costs across different $\alpha$ values.}
       \label{fig:tsp_correlation}
   \end{minipage}
   \begin{minipage}{0.48\textwidth}
       \centering
       \scriptsize
       \begin{tabular}{l>{\raggedright\arraybackslash}p{1.2cm}>{\raggedright\arraybackslash}p{1.0cm}>{\raggedright\arraybackslash}p{1.0cm}>{\raggedright\arraybackslash}p{1.0cm}}
           \toprule
           Method & \multicolumn{1}{c}{TSP} & \multicolumn{1}{c}{TSP-D$_1$} & \multicolumn{1}{c}{TSP-D$_2$} & \multicolumn{1}{c}{TSP-D$_3$} \\
           \midrule
           Concorde & \multicolumn{1}{r}{780.00} & \multicolumn{1}{r}{639.65} & \multicolumn{1}{r}{527.98} & \multicolumn{1}{r}{484.14} \\
           FI & \multicolumn{1}{r}{837.07} & \multicolumn{1}{r}{653.25} & \multicolumn{1}{r}{526.62} & \multicolumn{1}{r}{476.78} \\
           CI & \multicolumn{1}{r}{918.21} & \multicolumn{1}{r}{699.42} & \multicolumn{1}{r}{567.84} & \multicolumn{1}{r}{516.35} \\
           NN & \multicolumn{1}{r}{972.40} & \multicolumn{1}{r}{748.58} & \multicolumn{1}{r}{617.87} & \multicolumn{1}{r}{567.99} \\
           Random & \multicolumn{1}{r}{5297.03} & \multicolumn{1}{r}{1274.97} & \multicolumn{1}{r}{972.39} & \multicolumn{1}{r}{847.07} \\
           \bottomrule
       \end{tabular}
       \captionsetup{type=table}
       \caption{Average costs for initial TSP construction methods. Subscript denotes $\alpha$ values.}
       \label{tab:initial_tsp}
   \end{minipage}
\end{figure}

This section presents an empirical analysis for determining optimal configurations of the ICP algorithm. 
We first investigate various approaches for constructing the initial TSP tour for chain generation, presented in Figure \ref{fig:tsp_correlation} and Table \ref{tab:initial_tsp}. 
The analysis compares five methods: Concorde, an exact TSP solver that guarantees optimal solutions; Farthest Insertion (FI), a construction heuristic that iteratively inserts the node farthest from the current tour; Nearest Neighbor (NN), a greedy construction heuristic that sequentially adds the closest unvisited node; Closest Insertion (CI), a construction heuristic that iteratively inserts nodes with minimal insertion cost; and Random, which generates a random permutation of nodes.
We note that LKH-3 empirically achieves optimal solutions for this instance size, so we excluded it from the study.

\begin{table}[h]
   \centering
       \scriptsize
   \caption{Ablation study of chainlet construction methods for ICP and NICP across different $\alpha$ values on Set A unlimited instances. Times are in seconds.}
   \label{tab:chain_ablation}
   \begin{tabular}{
       >{\centering\arraybackslash}m{0.6cm}   %
       >{\raggedleft\arraybackslash}m{1.0cm}   %
       >{\raggedleft\arraybackslash}m{0.8cm}   %
       >{\raggedleft\arraybackslash}m{1.0cm}   %
       >{\raggedleft\arraybackslash}m{0.8cm}   %
       >{\raggedleft\arraybackslash}m{1.0cm}   %
       >{\raggedleft\arraybackslash}m{1.1cm}   %
       >{\raggedleft\arraybackslash}m{1.0cm}   %
       >{\raggedleft\arraybackslash}m{0.8cm}   %
       >{\raggedleft\arraybackslash}m{1.0cm}   %
       >{\raggedleft\arraybackslash}m{0.8cm}   %
       >{\raggedleft\arraybackslash}m{1.0cm}   %
       >{\raggedleft\arraybackslash}m{1.1cm}   %
   }
       \toprule
       & \multicolumn{6}{c}{Concorde} & \multicolumn{6}{c}{FI} \\
       \cmidrule(lr){2-7}\cmidrule(lr){8-13}
       & \multicolumn{2}{c}{ICP} & \multicolumn{2}{c}{NICP} & \multicolumn{2}{c}{Comp. (\%)} & \multicolumn{2}{c}{ICP} & \multicolumn{2}{c}{NICP} & \multicolumn{2}{c}{Comp. (\%)} \\
       \cmidrule(lr){2-3}\cmidrule(lr){4-5}\cmidrule(lr){6-7}\cmidrule(lr){8-9}\cmidrule(lr){10-11}\cmidrule(lr){12-13}
       $\alpha$ & Obj & Time & Obj & Time & Obj & Time & Obj & Time & Obj & Time & Obj & Time \\
       \midrule
       1 & 1475.36 & 1.93 & 1476.19 & 1.98 & 0.06\% & 2.6\% & 1523.03 & 0.76 & 1525.52 & 0.61 & 0.16\% & -19.7\% \\
       2 & 1193.76 & 4.56 & 1195.17 & 3.29 & 0.12\% & -27.9\% & 1210.07 & 4.02 & 1212.51 & 2.09 & 0.20\% & -48.0\% \\
       3 & 1076.03 & 6.98 & 1078.92 & 4.91 & 0.27\% & -29.7\% & 1068.15 & 7.24 & 1070.53 & 3.66 & 0.22\% & -49.4\% \\
       \bottomrule
   \end{tabular}
\end{table}

For this analysis, we generated 100 uniform instances with 100 nodes and executed ICP with each TSP construction method for chain initialization. 
The results, averaged across all instances, demonstrate that the quality of the initial TSP tour significantly influences the final TSP-D solution quality. 
As shown in Figure \ref{fig:tsp_correlation}, where the Random method is excluded from the regression analysis, TSP and TSP-D costs demonstrate an apparent linear relationship. 
However, this correlation diminishes as $\alpha$ increases, evidenced by the decreasing R² values from 0.741 at $\alpha=1$ to 0.519 at $\alpha=3$.
This diminishing correlation motivates an ablation study presented in Table \ref{tab:chain_ablation}, where we compare ICP and NICP performance using Concorde versus FI initialization across different $\alpha$ values on Set A unlimited instances.
As shown, the performance of Concorde initialization is strong, but it tends to degrade as $\alpha$ increases, which matches our analysis. 
Nevertheless, given its strong overall performance, we selected Concorde as the default configuration.

\begin{table}[h]
   \centering
       \scriptsize
   \caption{Performance comparison of chainlet input tour construction methods with varying maximum node sizes. Results are averaged over 100 uniform instances with 100 nodes. Times are in seconds.}
   \label{tab:chainlet_hyper} 
   \vspace{1em}
   \begin{tabular}{
       >{\centering\arraybackslash}m{0.4cm} %
       >{\centering\arraybackslash}m{0.5cm} %
       >{\raggedleft\arraybackslash}m{0.8cm} >{\raggedleft\arraybackslash}m{0.6cm} %
       >{\raggedleft\arraybackslash}m{0.8cm} >{\raggedleft\arraybackslash}m{0.6cm} %
       >{\raggedleft\arraybackslash}m{0.8cm} >{\raggedleft\arraybackslash}m{0.6cm} %
       >{\raggedleft\arraybackslash}m{0.8cm} >{\raggedleft\arraybackslash}m{0.6cm} %
       >{\raggedleft\arraybackslash}m{0.8cm} >{\raggedleft\arraybackslash}m{0.6cm} %
   }
       \toprule
       & & \multicolumn{2}{c}{LKH-3} & \multicolumn{2}{c}{FI} & \multicolumn{2}{c}{NN} & \multicolumn{2}{c}{CI} & \multicolumn{2}{c}{Random} \\
       \cmidrule(lr){3-4} \cmidrule(lr){5-6} \cmidrule(lr){7-8} \cmidrule(lr){9-10} \cmidrule(lr){11-12}
       $\alpha$ & Size & Obj & Time & Obj & Time & Obj & Time & Obj & Time & Obj & Time \\
       \midrule
       \multirow{6}{*}{$1$}
       & 17 & \multicolumn{1}{r}{645.78} & \multicolumn{1}{r}{0.18} & \multicolumn{1}{r}{645.10} & \multicolumn{1}{r}{0.10} & \multicolumn{1}{r}{645.41} & \multicolumn{1}{r}{0.15} & \multicolumn{1}{r}{645.86} & \multicolumn{1}{r}{0.12} & \multicolumn{1}{r}{646.85} & \multicolumn{1}{r}{0.41} \\
       & 18 & \multicolumn{1}{r}{644.73} & \multicolumn{1}{r}{0.13} & \multicolumn{1}{r}{644.18} & \multicolumn{1}{r}{0.08} & \multicolumn{1}{r}{644.56} & \multicolumn{1}{r}{0.11} & \multicolumn{1}{r}{644.54} & \multicolumn{1}{r}{0.10} & \multicolumn{1}{r}{644.22} & \multicolumn{1}{r}{0.40} \\
       & 19 & \multicolumn{1}{r}{643.89} & \multicolumn{1}{r}{0.15} & \multicolumn{1}{r}{643.07} & \multicolumn{1}{r}{0.11} & \multicolumn{1}{r}{643.39} & \multicolumn{1}{r}{0.16} & \multicolumn{1}{r}{643.66} & \multicolumn{1}{r}{0.13} & \multicolumn{1}{r}{642.51} & \multicolumn{1}{r}{0.48} \\
       & 20 & \multicolumn{1}{r}{643.15} & \multicolumn{1}{r}{0.18} & \multicolumn{1}{r}{642.38} & \multicolumn{1}{r}{0.13} & \multicolumn{1}{r}{642.61} & \multicolumn{1}{r}{0.20} & \multicolumn{1}{r}{642.74} & \multicolumn{1}{r}{0.17} & \multicolumn{1}{r}{641.64} & \multicolumn{1}{r}{0.61} \\
       & 21 & \multicolumn{1}{r}{642.53} & \multicolumn{1}{r}{0.22} & \multicolumn{1}{r}{641.71} & \multicolumn{1}{r}{0.16} & \multicolumn{1}{r}{642.00} & \multicolumn{1}{r}{0.24} & \multicolumn{1}{r}{642.23} & \multicolumn{1}{r}{0.20} & \multicolumn{1}{r}{641.02} & \multicolumn{1}{r}{0.72} \\
       & 22 & \multicolumn{1}{r}{641.84} & \multicolumn{1}{r}{0.26} & \multicolumn{1}{r}{640.76} & \multicolumn{1}{r}{0.20} & \multicolumn{1}{r}{641.64} & \multicolumn{1}{r}{0.29} & \multicolumn{1}{r}{641.56} & \multicolumn{1}{r}{0.26} & \multicolumn{1}{r}{640.53} & \multicolumn{1}{r}{0.92} \\
       \midrule
       \multirow{6}{*}{$2$}
       & 17 & \multicolumn{1}{r}{530.94} & \multicolumn{1}{r}{1.08} & \multicolumn{1}{r}{530.10} & \multicolumn{1}{r}{0.75} & \multicolumn{1}{r}{530.38} & \multicolumn{1}{r}{0.96} & \multicolumn{1}{r}{529.30} & \multicolumn{1}{r}{0.84} & \multicolumn{1}{r}{528.22} & \multicolumn{1}{r}{2.16} \\
       & 18 & \multicolumn{1}{r}{530.34} & \multicolumn{1}{r}{0.46} & \multicolumn{1}{r}{529.34} & \multicolumn{1}{r}{0.34} & \multicolumn{1}{r}{529.17} & \multicolumn{1}{r}{0.43} & \multicolumn{1}{r}{528.78} & \multicolumn{1}{r}{0.36} & \multicolumn{1}{r}{526.82} & \multicolumn{1}{r}{1.02} \\
       & 19 & \multicolumn{1}{r}{529.43} & \multicolumn{1}{r}{0.56} & \multicolumn{1}{r}{528.28} & \multicolumn{1}{r}{0.43} & \multicolumn{1}{r}{528.83} & \multicolumn{1}{r}{0.51} & \multicolumn{1}{r}{527.88} & \multicolumn{1}{r}{0.47} & \multicolumn{1}{r}{525.64} & \multicolumn{1}{r}{1.23} \\
       & 20 & \multicolumn{1}{r}{529.02} & \multicolumn{1}{r}{0.63} & \multicolumn{1}{r}{527.88} & \multicolumn{1}{r}{0.50} & \multicolumn{1}{r}{528.19} & \multicolumn{1}{r}{0.65} & \multicolumn{1}{r}{527.53} & \multicolumn{1}{r}{0.55} & \multicolumn{1}{r}{524.89} & \multicolumn{1}{r}{1.40} \\
       & 21 & \multicolumn{1}{r}{528.55} & \multicolumn{1}{r}{0.74} & \multicolumn{1}{r}{527.25} & \multicolumn{1}{r}{0.60} & \multicolumn{1}{r}{527.41} & \multicolumn{1}{r}{0.78} & \multicolumn{1}{r}{526.71} & \multicolumn{1}{r}{0.69} & \multicolumn{1}{r}{524.41} & \multicolumn{1}{r}{1.79} \\
       & 22 & \multicolumn{1}{r}{528.23} & \multicolumn{1}{r}{0.87} & \multicolumn{1}{r}{526.51} & \multicolumn{1}{r}{0.78} & \multicolumn{1}{r}{526.94} & \multicolumn{1}{r}{0.95} & \multicolumn{1}{r}{526.41} & \multicolumn{1}{r}{0.84} & \multicolumn{1}{r}{523.86} & \multicolumn{1}{r}{2.23} \\
       \midrule
       \multirow{6}{*}{$3$}
       & 17 & \multicolumn{1}{r}{488.69} & \multicolumn{1}{r}{1.66} & \multicolumn{1}{r}{487.82} & \multicolumn{1}{r}{1.27} & \multicolumn{1}{r}{486.44} & \multicolumn{1}{r}{1.54} & \multicolumn{1}{r}{486.20} & \multicolumn{1}{r}{1.37} & \multicolumn{1}{r}{481.45} & \multicolumn{1}{r}{3.18} \\
       & 18 & \multicolumn{1}{r}{487.47} & \multicolumn{1}{r}{0.74} & \multicolumn{1}{r}{486.76} & \multicolumn{1}{r}{0.57} & \multicolumn{1}{r}{485.36} & \multicolumn{1}{r}{0.68} & \multicolumn{1}{r}{484.56} & \multicolumn{1}{r}{0.61} & \multicolumn{1}{r}{478.65} & \multicolumn{1}{r}{1.50} \\
       & 19 & \multicolumn{1}{r}{486.75} & \multicolumn{1}{r}{0.86} & \multicolumn{1}{r}{484.90} & \multicolumn{1}{r}{0.76} & \multicolumn{1}{r}{484.86} & \multicolumn{1}{r}{0.79} & \multicolumn{1}{r}{483.54} & \multicolumn{1}{r}{0.72} & \multicolumn{1}{r}{477.30} & \multicolumn{1}{r}{1.70} \\
       & 20 & \multicolumn{1}{r}{485.16} & \multicolumn{1}{r}{1.07} & \multicolumn{1}{r}{483.07} & \multicolumn{1}{r}{0.93} & \multicolumn{1}{r}{484.33} & \multicolumn{1}{r}{0.97} & \multicolumn{1}{r}{482.97} & \multicolumn{1}{r}{0.89} & \multicolumn{1}{r}{476.52} & \multicolumn{1}{r}{1.92} \\
       & 21 & \multicolumn{1}{r}{484.63} & \multicolumn{1}{r}{1.18} & \multicolumn{1}{r}{482.12} & \multicolumn{1}{r}{1.07} & \multicolumn{1}{r}{483.59} & \multicolumn{1}{r}{1.25} & \multicolumn{1}{r}{482.29} & \multicolumn{1}{r}{1.09} & \multicolumn{1}{r}{475.18} & \multicolumn{1}{r}{2.47} \\
       & 22 & \multicolumn{1}{r}{483.95} & \multicolumn{1}{r}{1.48} & \multicolumn{1}{r}{481.06} & \multicolumn{1}{r}{1.29} & \multicolumn{1}{r}{482.89} & \multicolumn{1}{r}{1.50} & \multicolumn{1}{r}{481.94} & \multicolumn{1}{r}{1.25} & \multicolumn{1}{r}{474.79} & \multicolumn{1}{r}{2.73} \\
       \bottomrule
   \end{tabular}
\end{table}

\begin{table}[h]
   \centering
       \scriptsize
   \caption{Ablation study of chainlet construction methods and maximum node sizes for ICP and NICP across different $\alpha$ values on Set A unlimited instances. Times are in seconds.}
   \label{tab:chainlet_ablation}
   \begin{tabular}{
       >{\centering\arraybackslash}m{0.6cm}   %
       >{\raggedleft\arraybackslash}m{1.0cm}   %
       >{\raggedleft\arraybackslash}m{0.8cm}   %
       >{\raggedleft\arraybackslash}m{1.0cm}   %
       >{\raggedleft\arraybackslash}m{0.8cm}   %
       >{\raggedleft\arraybackslash}m{1.0cm}   %
       >{\raggedleft\arraybackslash}m{0.8cm}   %
       >{\raggedleft\arraybackslash}m{1.0cm}   %
       >{\raggedleft\arraybackslash}m{0.8cm}   %
       >{\raggedleft\arraybackslash}m{1.0cm}   %
       >{\raggedleft\arraybackslash}m{0.8cm}   %
       >{\raggedleft\arraybackslash}m{1.0cm}   %
       >{\raggedleft\arraybackslash}m{0.8cm}   %
   }
       \toprule
       & \multicolumn{4}{c}{Baseline} & \multicolumn{4}{c}{Random} & \multicolumn{4}{c}{Max node 25} \\
       \cmidrule(lr){2-5}\cmidrule(lr){6-9}\cmidrule(lr){10-13}
       & \multicolumn{2}{c}{ICP} & \multicolumn{2}{c}{NICP} & \multicolumn{2}{c}{ICP} & \multicolumn{2}{c}{NICP} & \multicolumn{2}{c}{ICP} & \multicolumn{2}{c}{NICP} \\
       \cmidrule(lr){2-3}\cmidrule(lr){4-5}\cmidrule(lr){6-7}\cmidrule(lr){8-9}\cmidrule(lr){10-11}\cmidrule(lr){12-13}
       $\alpha$ & Obj & Time & Obj & Time & Obj & Time & Obj & Time & Obj & Time & Obj & Time \\
       \midrule
       1 & 1475.36 & 1.93 & 1476.19 & 1.98 & 1479.22 & 4.03 & 1479.68 & 3.37 & 1467.81 & 3.29 & 1469.01 & 2.89 \\
       2 & 1193.76 & 4.56 & 1195.17 & 3.29 & 1188.45 & 13.11 & 1191.65 & 7.29 & 1191.79 & 12.23 & 1194.73 & 7.00 \\
       3 & 1076.03 & 6.98 & 1078.92 & 4.91 & 1067.15 & 17.81 & 1069.28 & 9.87 & 1064.88 & 20.92 & 1066.44 & 11.55 \\
       \bottomrule
   \end{tabular}
\end{table}

We then analyze methods for constructing chainlet input paths, with results presented in Table \ref{tab:chainlet_hyper}. 
For this analysis, we used LKH-3 instead of Concorde, as LKH-3 empirically achieves optimal solutions for instances of chainlet size while requiring less computational time. 
The other methods (FI, NN, CI, and Random) remain identical to those used in the initial chain construction analysis but are slightly modified to avoid altering the chainlet's end node.

For the experiment, we again generated 100 uniform instances with 100 nodes. 
Random method achieves superior objective values as $\alpha$ increases, but requires substantially longer computation times and introduces non-deterministic behavior. 
Other methods exhibit minimal differences in solution quality; therefore, FI was selected based on its computational efficiency.
Analysis of maximum node size configurations demonstrates that smaller maximum node sizes necessitate additional iterations for convergence, while larger sizes increase per-iteration computational burden. 
Computational time initially decreases, then increases as the size grows. 
Correspondingly, larger maximum node sizes produce superior objective values. 
Balancing these considerations, we selected a maximum node size of 20 as the default configuration.
In an ablation study presented in Table \ref{tab:chainlet_ablation}, the Random method achieves better objective values for $\alpha=2, 3$, but incurs longer computation times.
The maximum node size of 25 provides marginal improvements in solution quality but substantially increases computation time, supporting our selection of the default configuration.

\newpage
\section{Neural Network Training and Performance Evaluation} \label{sec:training_and_performance}

\begin{figure}[!ht]
    \centering
    \begin{subfigure}{0.7\textwidth}
        \centering
        \includegraphics[width=\textwidth]{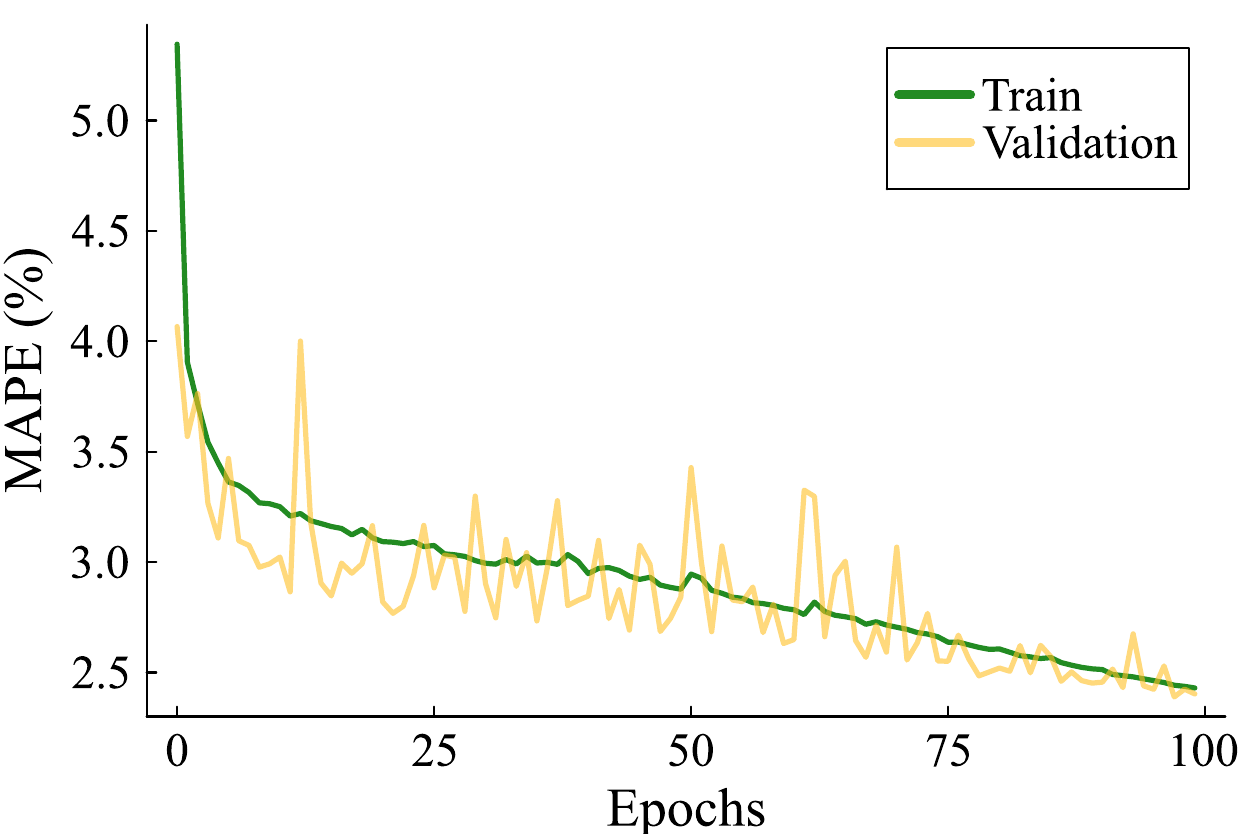}
        \caption{Base model training progress}
        \label{fig:training_base}
    \end{subfigure}
    
    \vspace{1em} %
    
    \begin{subfigure}{0.7\textwidth}
        \centering
        \includegraphics[width=\textwidth]{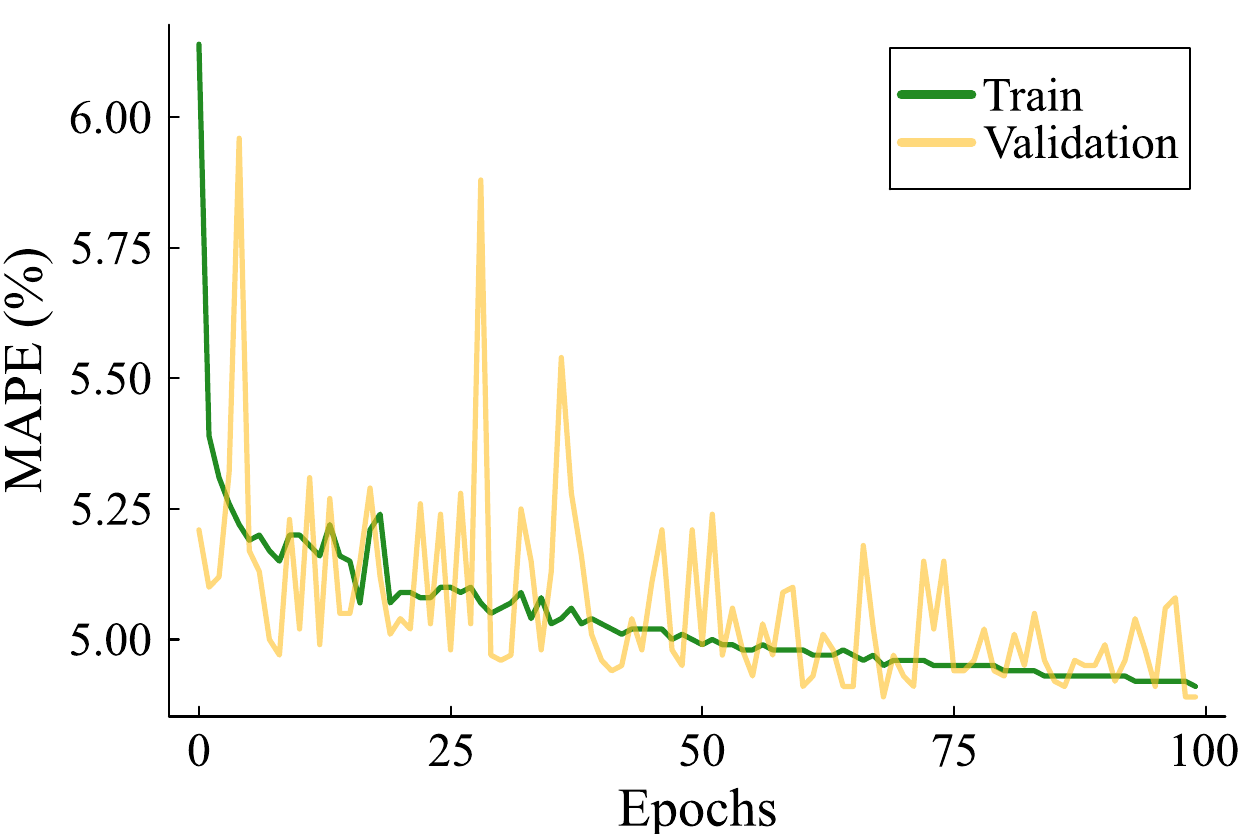}
        \caption{Transfer learning training progress}
        \label{fig:training_transfer}
    \end{subfigure}
    \caption{Training progress for base model and transfer learning extension.}
    \label{fig:training_progress}
\end{figure}

\begin{figure}[!ht]
    \centering
        \begin{subfigure}{0.45\textwidth}
        \centering
                \includegraphics[width=\textwidth]{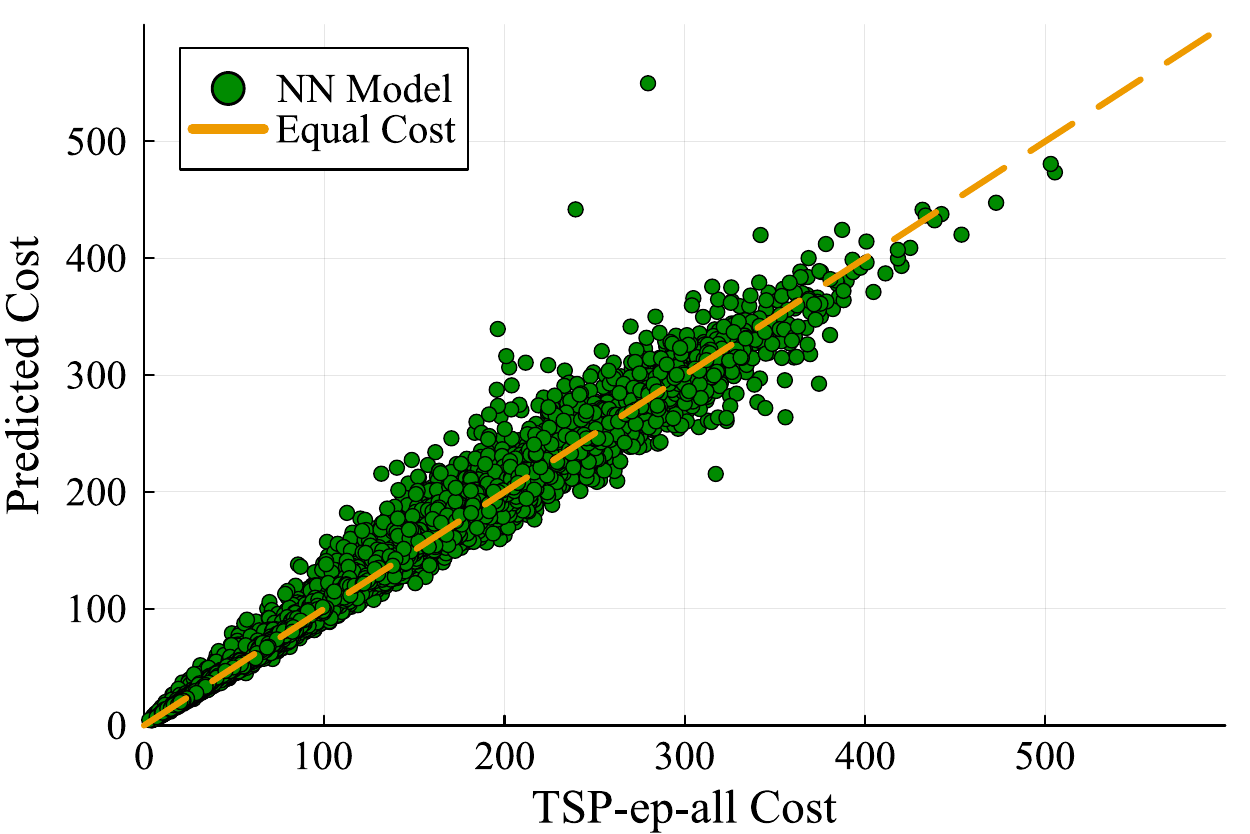}
        \caption{Scatter plot for 1-center}
        \label{fig:scatter_1center}
    \end{subfigure}
    \hfill
    \begin{subfigure}{0.45\textwidth}
        \centering
                \includegraphics[width=\textwidth]{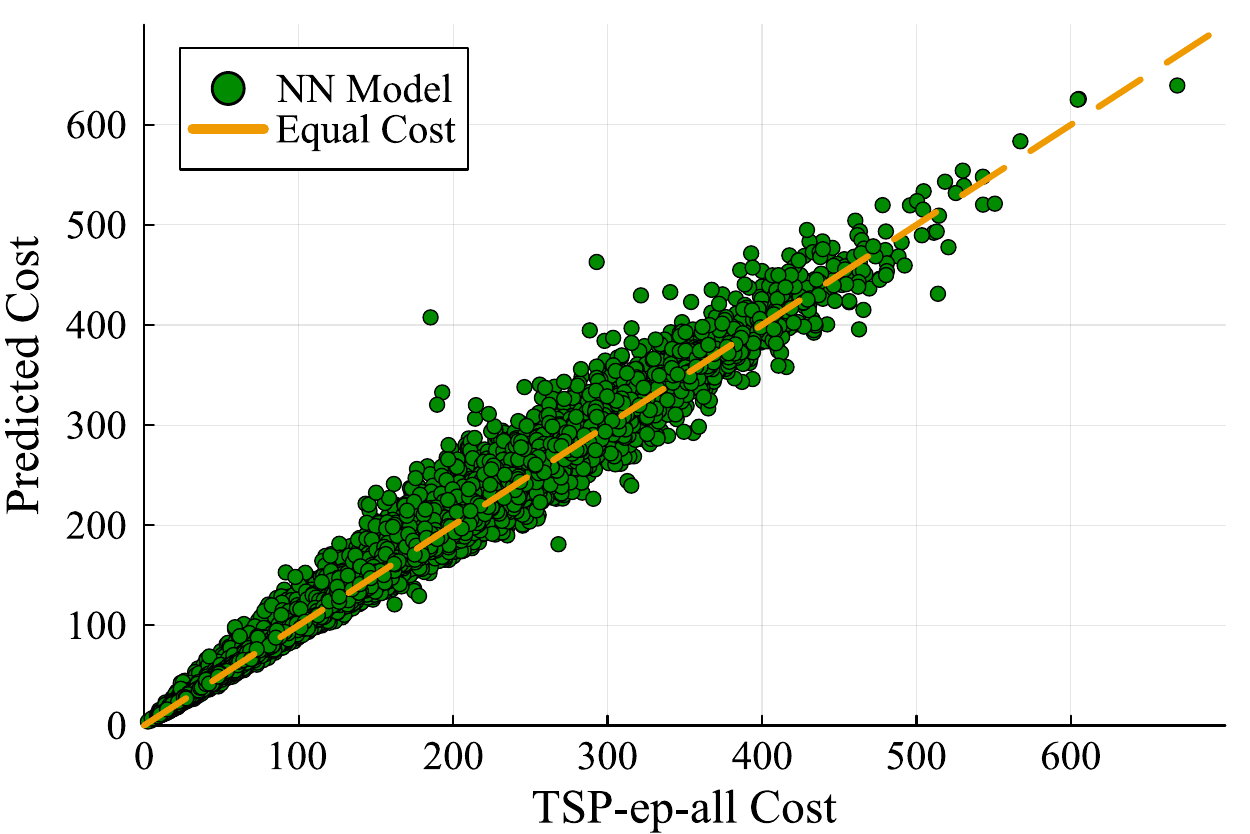}
        \caption{Scatter plot for 2-center}
        \label{fig:scatter_2center}
    \end{subfigure}
    \vspace{1em}
    
        \begin{subfigure}{0.45\textwidth}
        \centering
                \includegraphics[width=\textwidth]{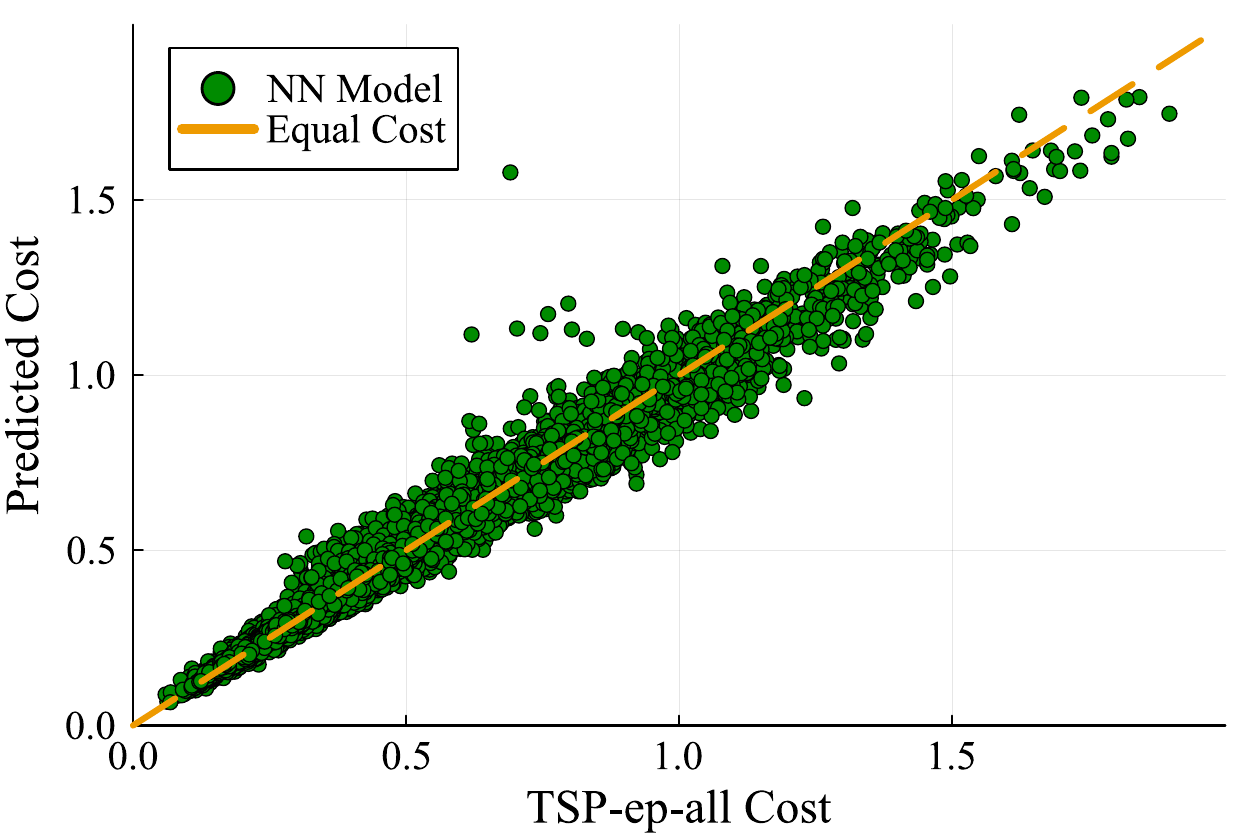}
        \caption{Scatter plot for Amsterdam}
        \label{fig:scatter_amsterdam}
    \end{subfigure}
    \hfill
        \begin{subfigure}{0.45\textwidth}
        \centering
                \includegraphics[width=\textwidth]{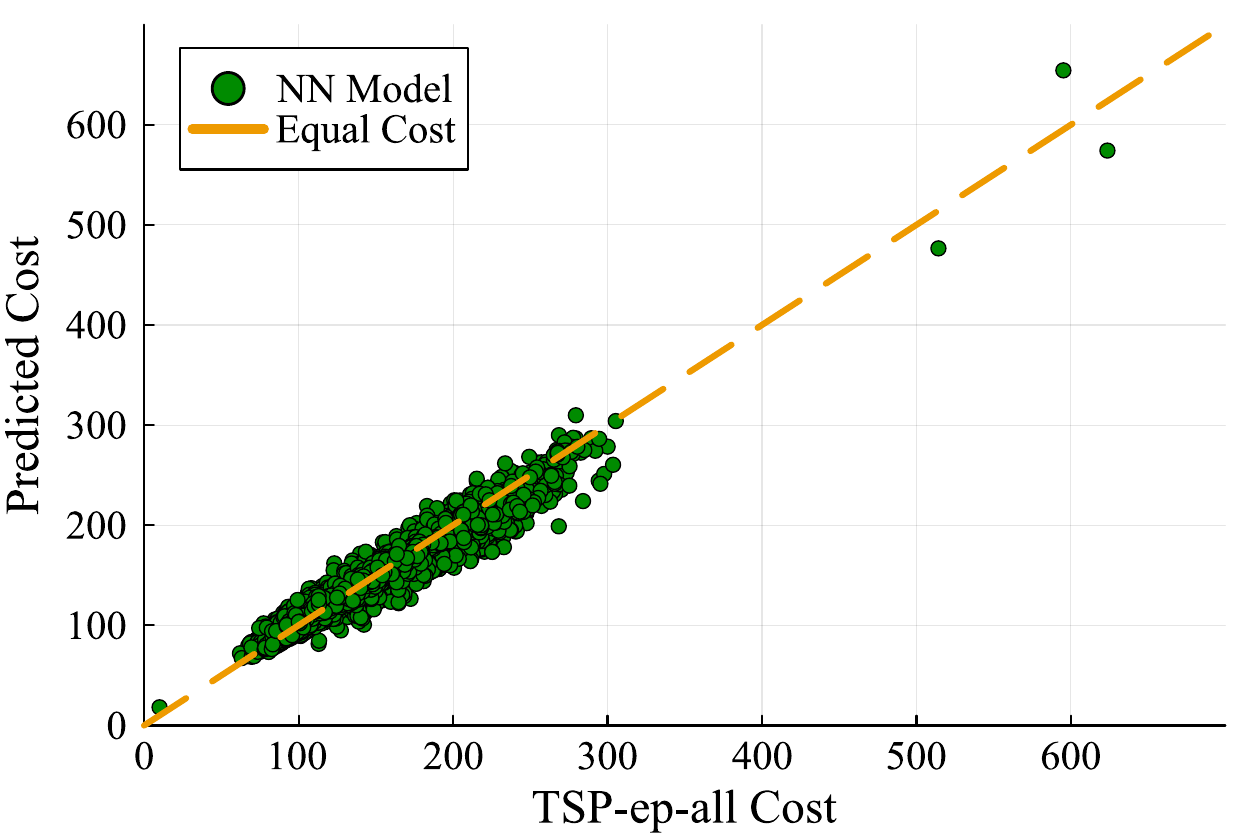}
    \caption{Scatter plot for flying range extension}
        \label{fig:scatter_flying_range}
    \end{subfigure}
    \vspace{1em}

    \caption{Predicted vs. actual costs for test sets.}
    \label{fig:ood_combined}
\end{figure}

This section presents supplementary figures supporting the analysis of the neural network's performance. 
Figure~\ref{fig:training_progress} illustrates the training progress for both the base model and the transfer learning extension, demonstrating stable convergence.
Complementing the MAPE results in Table~\ref{tab:nn_test}, Figure~\ref{fig:ood_combined} visualizes the predictor's high accuracy via scatter plots. 
These plots confirm the model's robust generalization to OOD data and its successful extension to the flying range constraint via transfer learning.

\newpage
\begin{landscape}

\section{Detailed Results of Computational Experiments} \label{sec:detailed_results}

\begin{table}[h]
    \centering
    \scriptsize
    \caption{(Set~A, uniform) Results of ICP and NICP for Agatz et al. (2018) uniform instances. 
        Comparison (\%) represents the relative difference of NICP compared to ICP.
    Times are in seconds.}
    \label{tab:SetA_uniform}
    \vspace{1em}
    \begin{tabular}{
        >{\centering\arraybackslash}m{0.6cm} %
        >{\centering\arraybackslash}m{0.6cm} %
        >{\raggedleft\arraybackslash}m{1.0cm} >{\raggedleft\arraybackslash}m{0.8cm} %
        >{\raggedleft\arraybackslash}m{1.0cm} >{\raggedleft\arraybackslash}m{1.0cm} >{\raggedleft\arraybackslash}m{1.0cm} >{\raggedleft\arraybackslash}m{0.8cm} %
        >{\raggedleft\arraybackslash}m{1.0cm} >{\raggedleft\arraybackslash}m{0.8cm} %
        >{\raggedleft\arraybackslash}m{0.8cm} >{\raggedleft\arraybackslash}m{0.8cm} %
        >{\raggedleft\arraybackslash}m{1.0cm} >{\raggedleft\arraybackslash}m{0.8cm} %
        >{\raggedleft\arraybackslash}m{0.8cm} >{\raggedleft\arraybackslash}m{0.8cm} %
        >{\raggedleft\arraybackslash}m{0.8cm} >{\raggedleft\arraybackslash}m{0.8cm} %
    }
        \toprule
        & & \multicolumn{2}{c}{DPS$_{25}$} & \multicolumn{4}{c}{HGA-TAC$^+$} & \multicolumn{2}{c}{ICP} & \multicolumn{2}{c}{Gap Over (\%)} & \multicolumn{2}{c}{NICP} & \multicolumn{2}{c}{Gap Over (\%)} & \multicolumn{2}{c}{Comparison (\%)} \\
        \cmidrule(lr){3-4} \cmidrule(lr){5-8} \cmidrule(lr){9-10} \cmidrule(lr){11-12} \cmidrule(lr){13-14} \cmidrule(lr){15-16} \cmidrule(lr){17-18}
         & $N$ & Obj & Time & Best & Mean & Worst & Time & Obj & Time & DPS & HGA & Obj & Time & DPS & HGA & Obj & Time \\
               \midrule
        \multirow{7}{*}{\centering {$\alpha=1$}} 
& 50 & 498.18 & 0.09 & 490.72 & 496.29 & 500.56 & 4.46 & 494.46 & 0.12 & -0.75 & -0.37 & 494.60 & 0.12 & -0.72 & -0.34 & 0.03 & 3.50 \\
& 75 & 574.21 & 0.14 & 570.60 & 574.61 & 576.95 & 7.51 & 573.72 & 0.20 & -0.09 & -0.16 & 573.92 & 0.20 & -0.05 & -0.12 & 0.04 & 1.72 \\
& 100 & 655.94 & 0.18 & 651.49 & 655.83 & 659.26 & 11.75 & 649.31 & 0.28 & -1.01 & -0.99 & 650.99 & 0.26 & -0.75 & -0.74 & 0.26 & -5.65 \\
& 175 & 841.40 & 1.11 & 837.70 & 841.08 & 844.15 & 22.68 & 830.11 & 2.13 & -1.34 & -1.31 & 831.32 & 1.53 & -1.20 & -1.16 & 0.15 & -28.38 \\
& 250 & 997.01 & 1.46 & 993.19 & 997.10 & 999.89 & 46.83 & 984.38 & 1.46 & -1.27 & -1.28 & 985.70 & 1.40 & -1.13 & -1.14 & 0.13 & -4.07 \\
& 375 & 1194.42 & 4.30 & 1194.07 & 1197.97 & 1201.13 & 80.53 & 1180.41 & 4.84 & -1.17 & -1.47 & 1180.47 & 4.82 & -1.17 & -1.46 & 0.01 & -0.39 \\
& 500 & 1372.16 & 7.61 & 1373.86 & 1377.34 & 1380.49 & 135.97 & 1354.84 & 8.22 & -1.26 & -1.63 & 1355.42 & 8.39 & -1.22 & -1.59 & 0.04 & 2.14 \\
        \cmidrule(lr){2-18}
& Avg. & 911.21 & 2.21 & 908.21 & 912.27 & 915.39 & 47.34 & 901.09 & 2.53 & -1.11 & -1.23 & 901.86 & 2.46 & -1.03 & -1.14 & 0.09 & -2.63 \\
        \midrule
        \multirow{7}{*}{\centering {$\alpha=2$}} 
& 50 & 420.55 & 0.19 & 405.41 & 411.37 & 418.96 & 3.81 & 409.26 & 0.60 & -2.69 & -0.51 & 409.21 & 0.35 & -2.70 & -0.53 & -0.01 & -41.20 \\
& 75 & 482.77 & 0.25 & 470.73 & 479.39 & 489.43 & 6.87 & 473.57 & 0.90 & -1.91 & -1.21 & 473.60 & 0.52 & -1.90 & -1.21 & 0.01 & -41.97 \\
& 100 & 559.33 & 0.32 & 548.43 & 557.26 & 567.80 & 10.74 & 539.33 & 1.36 & -3.58 & -3.22 & 540.97 & 0.76 & -3.28 & -2.92 & 0.30 & -44.01 \\
& 175 & 712.08 & 1.85 & 703.03 & 712.52 & 721.59 & 32.15 & 693.49 & 3.19 & -2.61 & -2.67 & 693.65 & 2.66 & -2.59 & -2.65 & 0.02 & -16.61 \\
& 250 & 843.48 & 1.65 & 841.84 & 850.82 & 859.48 & 57.17 & 815.81 & 4.32 & -3.28 & -4.11 & 816.50 & 3.01 & -3.20 & -4.03 & 0.08 & -30.39 \\
& 375 & 1015.21 & 4.50 & 1021.42 & 1033.42 & 1047.58 & 126.95 & 985.48 & 9.58 & -2.93 & -4.64 & 985.86 & 5.93 & -2.89 & -4.60 & 0.04 & -38.11 \\
& 500 & 1163.93 & 9.85 & 1171.22 & 1184.00 & 1199.13 & 212.27 & 1124.45 & 14.23 & -3.39 & -5.03 & 1126.30 & 11.62 & -3.23 & -4.87 & 0.16 & -18.31 \\
        \cmidrule(lr){2-18}
& Avg. & 774.98 & 2.73 & 771.13 & 780.78 & 791.53 & 68.82 & 751.44 & 5.20 & -3.04 & -3.76 & 752.11 & 3.72 & -2.95 & -3.67 & 0.09 & -28.45 \\
        \midrule
        \multirow{7}{*}{\centering {$\alpha=3$}} 
& 50 & 393.03 & 0.21 & 366.83 & 376.26 & 386.33 & 3.62 & 369.34 & 1.04 & -6.03 & -1.84 & 369.59 & 0.59 & -5.96 & -1.77 & 0.07 & -43.07 \\
& 75 & 454.40 & 0.28 & 428.44 & 440.67 & 455.01 & 7.10 & 429.81 & 1.78 & -5.41 & -2.46 & 430.80 & 0.95 & -5.19 & -2.24 & 0.23 & -46.59 \\
& 100 & 529.67 & 0.42 & 505.52 & 515.81 & 525.71 & 11.26 & 498.73 & 2.20 & -5.84 & -3.31 & 499.85 & 1.30 & -5.63 & -3.09 & 0.23 & -40.83 \\
& 175 & 670.31 & 1.76 & 645.08 & 658.70 & 674.24 & 33.57 & 638.14 & 4.87 & -4.80 & -3.12 & 639.64 & 3.38 & -4.58 & -2.89 & 0.23 & -30.63 \\
& 250 & 796.18 & 1.79 & 776.97 & 790.88 & 805.76 & 67.62 & 754.92 & 6.87 & -5.18 & -4.55 & 755.56 & 4.42 & -5.10 & -4.47 & 0.08 & -35.70 \\
& 375 & 953.60 & 5.19 & 948.41 & 962.80 & 983.02 & 167.34 & 906.71 & 13.23 & -4.92 & -5.83 & 910.35 & 8.50 & -4.53 & -5.45 & 0.40 & -35.78 \\
& 500 & 1096.91 & 9.55 & 1094.26 & 1108.14 & 1124.03 & 288.53 & 1037.74 & 20.04 & -5.39 & -6.35 & 1040.83 & 15.40 & -5.11 & -6.07 & 0.30 & -23.16 \\
        \cmidrule(lr){2-18}
& Avg. & 722.82 & 2.74 & 704.98 & 717.77 & 732.51 & 84.95 & 684.88 & 7.40 & -5.25 & -4.58 & 686.47 & 5.04 & -5.03 & -4.36 & 0.23 & -31.90 \\        \bottomrule
           \end{tabular}
\end{table}

\newpage
\begin{table}
    \centering
    \scriptsize
    \caption{(Set~A, 1-center) Results of ICP and NICP for Agatz et al. (2018) 1-center instances. 
        Comparison (\%) represents the relative difference of NICP compared to ICP.
    Times are in seconds.}
    \label{tab:SetA_singlecenter}
    \vspace{1em}
    \begin{tabular}{
        >{\centering\arraybackslash}m{0.6cm} %
        >{\centering\arraybackslash}m{0.6cm} %
        >{\raggedleft\arraybackslash}m{1.0cm} >{\raggedleft\arraybackslash}m{0.8cm} %
        >{\raggedleft\arraybackslash}m{1.0cm} >{\raggedleft\arraybackslash}m{1.0cm} >{\raggedleft\arraybackslash}m{1.0cm} >{\raggedleft\arraybackslash}m{0.8cm} %
        >{\raggedleft\arraybackslash}m{1.0cm} >{\raggedleft\arraybackslash}m{0.8cm} %
        >{\raggedleft\arraybackslash}m{0.8cm} >{\raggedleft\arraybackslash}m{0.8cm} %
        >{\raggedleft\arraybackslash}m{1.0cm} >{\raggedleft\arraybackslash}m{0.8cm} %
        >{\raggedleft\arraybackslash}m{0.8cm} >{\raggedleft\arraybackslash}m{0.8cm} %
        >{\raggedleft\arraybackslash}m{0.8cm} >{\raggedleft\arraybackslash}m{0.8cm} %
    }
        \toprule
        & & \multicolumn{2}{c}{DPS$_{25}$} & \multicolumn{4}{c}{HGA-TAC$^+$} & \multicolumn{2}{c}{ICP} & \multicolumn{2}{c}{Gap Over (\%)} & \multicolumn{2}{c}{NICP} & \multicolumn{2}{c}{Gap Over (\%)} & \multicolumn{2}{c}{Comparison (\%)} \\
        \cmidrule(lr){3-4} \cmidrule(lr){5-8} \cmidrule(lr){9-10} \cmidrule(lr){11-12} \cmidrule(lr){13-14} \cmidrule(lr){15-16} \cmidrule(lr){17-18}
         & $N$ & Obj & Time & Best & Mean & Worst & Time & Obj & Time & DPS & HGA & Obj & Time & DPS & HGA & Obj & Time \\
        \midrule
        \multirow{7}{*}{\centering {$\alpha=1$}} 
& 50 & 657.79 & 0.10 & 648.81 & 652.53 & 656.22 & 6.21 & 654.15 & 0.10 & -0.55 & 0.25 & 654.07 & 0.11 & -0.57 & 0.24 & -0.01 & 2.07 \\
& 75 & 891.08 & 0.27 & 876.22 & 880.99 & 884.84 & 9.74 & 880.43 & 0.25 & -1.20 & -0.06 & 880.55 & 0.25 & -1.18 & -0.05 & 0.01 & -0.87 \\
& 100 & 1065.96 & 0.23 & 1050.95 & 1057.91 & 1063.56 & 14.82 & 1059.37 & 0.25 & -0.62 & 0.14 & 1058.69 & 0.23 & -0.68 & 0.07 & -0.06 & -7.02 \\
& 175 & 1420.90 & 0.94 & 1412.94 & 1417.87 & 1425.89 & 26.38 & 1405.92 & 0.70 & -1.05 & -0.84 & 1406.74 & 0.69 & -1.00 & -0.78 & 0.06 & -1.27 \\
& 250 & 1669.73 & 1.00 & 1646.38 & 1650.70 & 1655.12 & 51.80 & 1635.21 & 1.54 & -2.07 & -0.94 & 1635.60 & 1.46 & -2.04 & -0.91 & 0.02 & -5.51 \\
& 375 & 2065.11 & 3.31 & 2050.65 & 2055.27 & 2059.91 & 81.45 & 2024.04 & 2.86 & -1.99 & -1.52 & 2026.78 & 3.68 & -1.86 & -1.39 & 0.14 & 28.54 \\
& 500 & 2416.56 & 5.09 & 2419.75 & 2426.39 & 2434.00 & 118.36 & 2383.17 & 5.95 & -1.38 & -1.78 & 2384.52 & 6.78 & -1.33 & -1.73 & 0.06 & 13.94 \\
        \cmidrule(lr){2-18}
& Avg. & 1509.87 & 1.58 & 1496.71 & 1501.72 & 1506.98 & 46.80 & 1486.56 & 1.71 & -1.54 & -1.01 & 1487.29 & 1.92 & -1.50 & -0.96 & 0.05 & 12.38 \\
        \midrule
        \multirow{7}{*}{\centering {$\alpha=2$}} 
& 50 & 521.28 & 0.17 & 498.29 & 514.02 & 527.04 & 3.88 & 507.24 & 0.41 & -2.69 & -1.32 & 510.26 & 0.28 & -2.12 & -0.73 & 0.59 & -30.70 \\
& 75 & 715.92 & 0.31 & 691.83 & 707.94 & 725.97 & 7.75 & 686.93 & 0.94 & -4.05 & -2.97 & 687.35 & 0.62 & -3.99 & -2.91 & 0.06 & -33.80 \\
& 100 & 858.86 & 0.37 & 844.77 & 863.00 & 881.65 & 11.78 & 825.78 & 1.40 & -3.85 & -4.31 & 825.49 & 1.00 & -3.88 & -4.35 & -0.03 & -28.26 \\
& 175 & 1153.42 & 0.87 & 1153.79 & 1170.88 & 1187.33 & 29.81 & 1114.55 & 2.75 & -3.37 & -4.81 & 1119.62 & 1.75 & -2.93 & -4.38 & 0.45 & -36.55 \\
& 250 & 1365.37 & 1.31 & 1357.97 & 1380.45 & 1404.67 & 55.35 & 1314.69 & 3.81 & -3.71 & -4.76 & 1315.25 & 2.62 & -3.67 & -4.72 & 0.04 & -31.34 \\
& 375 & 1704.30 & 3.71 & 1709.46 & 1729.58 & 1755.16 & 121.40 & 1645.26 & 7.30 & -3.46 & -4.88 & 1647.84 & 5.79 & -3.31 & -4.73 & 0.16 & -20.63 \\
& 500 & 2002.71 & 5.70 & 2028.46 & 2057.92 & 2101.49 & 202.62 & 1930.43 & 12.46 & -3.61 & -6.20 & 1928.57 & 9.27 & -3.70 & -6.29 & -0.10 & -25.60 \\
        \cmidrule(lr){2-18}
& Avg. & 1234.40 & 1.82 & 1229.33 & 1249.53 & 1272.62 & 63.87 & 1190.20 & 4.26 & -3.58 & -4.75 & 1191.52 & 3.13 & -3.47 & -4.64 & 0.11 & -26.63 \\
        \midrule
        \multirow{7}{*}{\centering {$\alpha=3$}} 
& 50 & 470.82 & 0.21 & 432.16 & 446.90 & 465.93 & 3.93 & 444.96 & 0.79 & -5.49 & -0.44 & 443.74 & 0.47 & -5.75 & -0.71 & -0.27 & -40.68 \\
& 75 & 646.71 & 0.43 & 603.66 & 626.09 & 651.39 & 7.84 & 604.05 & 1.72 & -6.60 & -3.52 & 607.82 & 0.90 & -6.01 & -2.92 & 0.62 & -47.74 \\
& 100 & 778.04 & 0.48 & 736.43 & 763.07 & 787.71 & 13.69 & 727.54 & 2.47 & -6.49 & -4.66 & 730.19 & 1.86 & -6.15 & -4.31 & 0.36 & -24.84 \\
& 175 & 1044.40 & 1.11 & 1016.23 & 1045.34 & 1069.95 & 35.30 & 989.47 & 4.29 & -5.26 & -5.35 & 988.32 & 3.01 & -5.37 & -5.45 & -0.12 & -29.88 \\
& 250 & 1255.22 & 1.50 & 1213.70 & 1243.02 & 1272.20 & 67.76 & 1176.40 & 6.99 & -6.28 & -5.36 & 1179.00 & 4.56 & -6.07 & -5.15 & 0.22 & -34.77 \\
& 375 & 1581.52 & 3.85 & 1549.78 & 1579.30 & 1622.85 & 152.29 & 1489.10 & 11.94 & -5.84 & -5.71 & 1490.44 & 8.74 & -5.76 & -5.63 & 0.09 & -26.82 \\
& 500 & 1863.66 & 6.81 & 1845.83 & 1884.57 & 1943.77 & 271.86 & 1736.60 & 17.48 & -6.82 & -7.85 & 1739.43 & 14.23 & -6.67 & -7.70 & 0.16 & -18.58 \\
        \cmidrule(lr){2-18}
& Avg. & 1134.31 & 2.07 & 1098.98 & 1126.55 & 1158.95 & 81.11 & 1064.30 & 6.84 & -6.17 & -5.53 & 1065.97 & 4.98 & -6.03 & -5.38 & 0.16 & -27.21 \\        \bottomrule
    \end{tabular}
\end{table}

\newpage
\begin{table}
    \centering
    \scriptsize
    \caption{(Set~A, 2-center) Results of ICP and NICP for Agatz et al. (2018) 2-center instances. 
        Comparison (\%) represents the relative difference of NICP compared to ICP.
    Times are in seconds.}
    \label{tab:SetA_doublecenter}
    \vspace{1em}
    \begin{tabular}{
        >{\centering\arraybackslash}m{0.6cm} %
        >{\centering\arraybackslash}m{0.6cm} %
        >{\raggedleft\arraybackslash}m{1.0cm} >{\raggedleft\arraybackslash}m{0.8cm} %
        >{\raggedleft\arraybackslash}m{1.0cm} >{\raggedleft\arraybackslash}m{1.0cm} >{\raggedleft\arraybackslash}m{1.0cm} >{\raggedleft\arraybackslash}m{0.8cm} %
        >{\raggedleft\arraybackslash}m{1.0cm} >{\raggedleft\arraybackslash}m{0.8cm} %
        >{\raggedleft\arraybackslash}m{0.8cm} >{\raggedleft\arraybackslash}m{0.8cm} %
        >{\raggedleft\arraybackslash}m{1.0cm} >{\raggedleft\arraybackslash}m{0.8cm} %
        >{\raggedleft\arraybackslash}m{0.8cm} >{\raggedleft\arraybackslash}m{0.8cm} %
        >{\raggedleft\arraybackslash}m{0.8cm} >{\raggedleft\arraybackslash}m{0.8cm} %
    }
        \toprule
        & & \multicolumn{2}{c}{DPS$_{25}$} & \multicolumn{4}{c}{HGA-TAC$^+$} & \multicolumn{2}{c}{ICP} & \multicolumn{2}{c}{Gap Over (\%)} & \multicolumn{2}{c}{NICP} & \multicolumn{2}{c}{Gap Over (\%)} & \multicolumn{2}{c}{Comparison (\%)} \\
        \cmidrule(lr){3-4} \cmidrule(lr){5-8} \cmidrule(lr){9-10} \cmidrule(lr){11-12} \cmidrule(lr){13-14} \cmidrule(lr){15-16} \cmidrule(lr){17-18}
         & $N$ & Obj & Time & Best & Mean & Worst & Time & Obj & Time & DPS & HGA & Obj & Time & DPS & HGA & Obj & Time \\
        \midrule
        \multirow{7}{*}{\centering {$\alpha=1$}} 
& 50 & 1010.88 & 0.12 & 989.70 & 998.41 & 1006.26 & 5.80 & 1002.50 & 0.13 & -0.83 & 0.41 & 1002.50 & 0.14 & -0.83 & 0.41 & 0.00 & 12.27 \\
& 75 & 1243.47 & 0.18 & 1223.55 & 1235.53 & 1243.52 & 10.77 & 1240.30 & 0.17 & -0.25 & 0.39 & 1240.60 & 0.18 & -0.23 & 0.41 & 0.02 & 9.54 \\
& 100 & 1405.76 & 0.24 & 1378.14 & 1388.26 & 1395.67 & 14.55 & 1390.74 & 0.25 & -1.07 & 0.18 & 1391.64 & 0.25 & -1.00 & 0.24 & 0.06 & 0.82 \\
& 175 & 1909.59 & 0.73 & 1893.29 & 1904.83 & 1912.78 & 31.24 & 1896.14 & 0.66 & -0.70 & -0.46 & 1896.78 & 0.65 & -0.67 & -0.42 & 0.03 & -0.96 \\
& 250 & 2260.61 & 0.96 & 2248.08 & 2255.19 & 2261.69 & 49.15 & 2233.25 & 1.19 & -1.21 & -0.97 & 2235.44 & 1.19 & -1.11 & -0.88 & 0.10 & -0.26 \\
& 375 & 2838.30 & 2.04 & 2825.75 & 2833.39 & 2839.63 & 75.37 & 2797.94 & 2.43 & -1.42 & -1.25 & 2796.92 & 2.59 & -1.46 & -1.29 & -0.04 & 6.84 \\
& 500 & 3314.61 & 6.97 & 3300.15 & 3309.19 & 3314.75 & 112.94 & 3266.89 & 6.09 & -1.44 & -1.28 & 3270.36 & 5.86 & -1.34 & -1.17 & 0.11 & -3.71 \\
        \cmidrule(lr){2-18}
& Avg. & 2068.95 & 1.55 & 2052.03 & 2061.11 & 2068.08 & 45.13 & 2045.26 & 1.56 & -1.14 & -0.77 & 2046.24 & 1.56 & -1.10 & -0.72 & 0.05 & 0.21 \\
        \midrule
        \multirow{7}{*}{\centering {$\alpha=2$}} 
& 50 & 825.24 & 0.16 & 792.69 & 806.64 & 825.86 & 3.88 & 810.55 & 0.38 & -1.78 & 0.48 & 810.56 & 0.30 & -1.78 & 0.49 & 0.00 & -21.40 \\
& 75 & 1019.05 & 0.28 & 992.25 & 1009.70 & 1032.50 & 7.83 & 1000.69 & 0.86 & -1.80 & -0.89 & 997.39 & 0.58 & -2.12 & -1.22 & -0.33 & -33.33 \\
& 100 & 1137.05 & 0.37 & 1124.36 & 1146.94 & 1168.89 & 13.17 & 1105.85 & 1.21 & -2.74 & -3.58 & 1111.70 & 0.79 & -2.23 & -3.07 & 0.53 & -34.70 \\
& 175 & 1569.42 & 0.86 & 1560.76 & 1584.32 & 1613.76 & 31.78 & 1521.75 & 2.56 & -3.04 & -3.95 & 1524.98 & 1.82 & -2.83 & -3.75 & 0.21 & -29.01 \\
& 250 & 1851.96 & 1.39 & 1867.40 & 1896.19 & 1919.35 & 59.21 & 1799.99 & 3.78 & -2.81 & -5.07 & 1802.28 & 2.60 & -2.68 & -4.95 & 0.13 & -31.12 \\
& 375 & 2346.80 & 2.56 & 2353.13 & 2382.33 & 2424.42 & 120.67 & 2265.04 & 7.27 & -3.48 & -4.92 & 2269.81 & 4.79 & -3.28 & -4.72 & 0.21 & -34.14 \\
& 500 & 2734.33 & 6.45 & 2767.99 & 2800.05 & 2836.37 & 199.78 & 2651.54 & 12.48 & -3.03 & -5.30 & 2653.45 & 9.94 & -2.96 & -5.24 & 0.07 & -20.33 \\
        \cmidrule(lr){2-18}
& Avg. & 1699.35 & 1.73 & 1698.48 & 1723.25 & 1751.22 & 64.74 & 1650.17 & 4.19 & -2.89 & -4.24 & 1652.42 & 3.02 & -2.76 & -4.11 & 0.14 & -28.09 \\
        \midrule
        \multirow{7}{*}{\centering {$\alpha=3$}} 
& 50 & 760.37 & 0.22 & 718.34 & 733.41 & 754.71 & 3.75 & 731.92 & 0.80 & -3.74 & -0.20 & 735.90 & 0.56 & -3.22 & 0.34 & 0.54 & -30.25 \\
& 75 & 932.58 & 0.33 & 877.31 & 904.34 & 931.13 & 7.83 & 892.17 & 1.42 & -4.33 & -1.35 & 892.42 & 1.02 & -4.31 & -1.32 & 0.03 & -28.50 \\
& 100 & 1037.43 & 0.50 & 998.08 & 1025.06 & 1055.65 & 13.72 & 989.66 & 2.36 & -4.60 & -3.45 & 993.77 & 1.31 & -4.21 & -3.05 & 0.42 & -44.72 \\
& 175 & 1449.84 & 1.00 & 1390.67 & 1428.00 & 1465.15 & 39.95 & 1368.77 & 4.93 & -5.59 & -4.15 & 1371.35 & 2.86 & -5.41 & -3.97 & 0.19 & -41.97 \\
& 250 & 1701.37 & 1.63 & 1658.41 & 1700.19 & 1741.22 & 71.78 & 1598.44 & 6.70 & -6.05 & -5.98 & 1602.92 & 4.57 & -5.79 & -5.72 & 0.28 & -31.75 \\
& 375 & 2170.07 & 4.18 & 2141.77 & 2182.94 & 2228.84 & 156.10 & 2041.50 & 11.17 & -5.92 & -6.48 & 2052.93 & 7.86 & -5.40 & -5.96 & 0.56 & -29.65 \\
& 500 & 2536.50 & 6.67 & 2505.45 & 2560.01 & 2626.01 & 286.67 & 2385.32 & 17.57 & -5.96 & -6.82 & 2395.01 & 13.61 & -5.58 & -6.45 & 0.41 & -22.56 \\
        \cmidrule(lr){2-18}
& Avg. & 1566.38 & 2.12 & 1524.42 & 1560.39 & 1599.46 & 85.00 & 1478.91 & 6.68 & -5.58 & -5.22 & 1484.33 & 4.70 & -5.24 & -4.87 & 0.37 & -29.63 \\        \bottomrule
    \end{tabular}
\end{table}

\newpage

\newpage
\begin{table}[h]
    \centering
    \scriptsize
    \caption{(Set~A, limited) Results of ICP and NICP for Agatz et al. (2018)'s instances with limited flying ranges.  
        Comparison (\%) represents the relative difference of NICP compared to ICP.
    Times are in seconds.}
    \label{tab:SetA_uniform}
    \vspace{1em}
    \begin{tabular}{
        >{\centering\arraybackslash}m{0.6cm} % 
        >{\centering\arraybackslash}m{0.6cm} % 
        >{\raggedleft\arraybackslash}m{1.0cm} >{\raggedleft\arraybackslash}m{0.8cm} % 
        >{\raggedleft\arraybackslash}m{1.0cm} >{\raggedleft\arraybackslash}m{1.0cm} >{\raggedleft\arraybackslash}m{1.0cm} >{\raggedleft\arraybackslash}m{0.8cm} % 
        >{\raggedleft\arraybackslash}m{1.0cm} >{\raggedleft\arraybackslash}m{0.8cm} % 
        >{\raggedleft\arraybackslash}m{0.8cm} >{\raggedleft\arraybackslash}m{0.8cm} % 
        >{\raggedleft\arraybackslash}m{1.0cm} >{\raggedleft\arraybackslash}m{0.8cm} % 
        >{\raggedleft\arraybackslash}m{0.8cm} >{\raggedleft\arraybackslash}m{0.8cm} % 
        >{\raggedleft\arraybackslash}m{0.8cm} >{\raggedleft\arraybackslash}m{0.8cm} % 
    }
        \toprule
        & & \multicolumn{2}{c}{DPS$_{25}$} & \multicolumn{4}{c}{HGA-TAC$^+$} & \multicolumn{2}{c}{ICP} & \multicolumn{2}{c}{Gap Over (\%)} & \multicolumn{2}{c}{NICP} & \multicolumn{2}{c}{Gap Over (\%)} & \multicolumn{2}{c}{Comparison (\%)} \\
        \cmidrule(lr){3-4} \cmidrule(lr){5-8} \cmidrule(lr){9-10} \cmidrule(lr){11-12} \cmidrule(lr){13-14} \cmidrule(lr){15-16} \cmidrule(lr){17-18}
        $N$ & $f$ & Obj & Time & Best & Mean & Worst & Time & Obj & Time & DPS & HGA & Obj & Time & DPS & HGA & Obj & Time \\
               \midrule
        \multirow{7}{*}{\centering {50}} 
& 5 & 595.57 & 0.06 & 595.62 & 595.62 & 595.62 & 7.65 & 595.57 & 0.29 & 0.00 & -0.01 & 595.57 & 0.35 & 0.00 & -0.01 & 0.00 & 23.99 \\
& 10 & 587.49 & 0.08 & 589.65 & 589.67 & 589.73 & 5.43 & 587.09 & 0.34 & -0.07 & -0.44 & 587.31 & 0.36 & -0.03 & -0.40 & 0.04 & 7.46 \\
& 15 & 564.59 & 0.09 & 568.15 & 568.43 & 568.61 & 3.94 & 561.11 & 0.43 & -0.62 & -1.29 & 561.11 & 0.37 & -0.62 & -1.29 & 0.00 & -13.21 \\
& 20 & 516.89 & 0.11 & 528.35 & 528.63 & 528.73 & 2.91 & 515.45 & 0.36 & -0.28 & -2.49 & 515.61 & 0.35 & -0.25 & -2.46 & 0.03 & -2.95 \\
& 30 & 459.13 & 0.14 & 465.98 & 470.44 & 475.47 & 4.66 & 450.30 & 0.58 & -1.92 & -4.28 & 450.70 & 0.43 & -1.84 & -4.20 & 0.09 & -26.29 \\
& 40 & 431.04 & 0.16 & 418.21 & 424.38 & 433.21 & 6.72 & 418.49 & 0.64 & -2.91 & -1.39 & 419.41 & 0.39 & -2.70 & -1.17 & 0.22 & -39.15 \\
& 50 & 427.33 & 0.17 & 408.76 & 413.65 & 420.10 & 7.00 & 410.75 & 0.75 & -3.88 & -0.70 & 411.86 & 0.44 & -3.62 & -0.43 & 0.27 & -41.69 \\
        \cmidrule(lr){2-18}
& Avg. & 511.72 & 0.11 & 510.67 & 512.97 & 515.92 & 5.47 & 505.54 & 0.48 & -1.21 & -1.45 & 505.94 & 0.38 & -1.13 & -1.37 & 0.08 & -20.40 \\
        \midrule
        \multirow{7}{*}{\centering {75}} 
& 5 & 692.43 & 0.10 & 692.82 & 692.83 & 692.84 & 14.89 & 692.43 & 0.53 & 0.00 & -0.06 & 692.43 & 0.60 & 0.00 & -0.06 & 0.00 & 12.51 \\
& 10 & 664.73 & 0.13 & 669.22 & 669.22 & 669.22 & 6.24 & 663.54 & 0.62 & -0.18 & -0.85 & 663.52 & 0.54 & -0.18 & -0.85 & -0.00 & -11.95 \\
& 15 & 612.91 & 0.18 & 625.19 & 625.19 & 625.19 & 4.56 & 607.54 & 0.68 & -0.88 & -2.82 & 607.61 & 0.61 & -0.86 & -2.81 & 0.01 & -10.59 \\
& 20 & 567.50 & 0.22 & 586.88 & 586.88 & 586.88 & 4.01 & 558.26 & 0.89 & -1.63 & -4.88 & 558.64 & 0.66 & -1.56 & -4.81 & 0.07 & -25.74 \\
& 30 & 505.39 & 0.25 & 500.78 & 513.56 & 524.21 & 11.52 & 493.65 & 1.09 & -2.32 & -3.88 & 493.20 & 0.68 & -2.41 & -3.96 & -0.09 & -36.97 \\
& 40 & 489.49 & 0.27 & 479.30 & 485.67 & 494.27 & 14.43 & 481.04 & 1.19 & -1.73 & -0.96 & 480.10 & 0.79 & -1.92 & -1.15 & -0.19 & -33.63 \\
& 50 & 483.76 & 0.29 & 470.39 & 477.43 & 486.37 & 13.34 & 475.99 & 1.19 & -1.61 & -0.30 & 477.11 & 0.76 & -1.37 & -0.07 & 0.24 & -36.38 \\
        \cmidrule(lr){2-18}
& Avg. & 573.74 & 0.21 & 574.94 & 578.68 & 582.71 & 9.85 & 567.49 & 0.88 & -1.09 & -1.93 & 567.51 & 0.66 & -1.09 & -1.93 & 0.00 & -24.95 \\
        \midrule
        \multirow{7}{*}{\centering {100}} 
& 5 & 780.43 & 0.14 & 780.59 & 780.59 & 780.59 & 23.73 & 780.32 & 0.77 & -0.01 & -0.03 & 780.32 & 0.87 & -0.01 & -0.03 & 0.00 & 11.96 \\
& 10 & 731.12 & 0.18 & 736.78 & 736.78 & 736.78 & 7.19 & 728.89 & 0.70 & -0.31 & -1.07 & 728.80 & 0.71 & -0.32 & -1.08 & -0.01 & 0.62 \\
& 15 & 660.37 & 0.25 & 675.69 & 675.69 & 675.69 & 5.51 & 653.40 & 1.09 & -1.05 & -3.30 & 654.35 & 0.86 & -0.91 & -3.16 & 0.15 & -21.54 \\
& 30 & 567.62 & 0.36 & 560.02 & 570.34 & 578.15 & 17.92 & 551.78 & 1.64 & -2.79 & -3.25 & 552.31 & 1.00 & -2.70 & -3.16 & 0.10 & -39.16 \\
& 50 & 559.72 & 0.37 & 546.40 & 554.48 & 563.11 & 20.92 & 539.64 & 1.70 & -3.59 & -2.68 & 540.62 & 1.01 & -3.41 & -2.50 & 0.18 & -40.55 \\
        \cmidrule(lr){2-18}
& Avg. & 659.85 & 0.26 & 659.89 & 663.57 & 666.86 & 15.05 & 650.80 & 1.18 & -1.37 & -1.92 & 651.28 & 0.89 & -1.30 & -1.85 & 0.07 & -24.87 \\        \bottomrule
           \end{tabular}
\end{table}

\newpage
\begin{table}
    \scriptsize
    \centering
    \caption{
     (Set~B) Results of ICP and NICP for \citet{bogyrbayeva2023deep} instances. 
    Average cost and time values are reported on 100 problem instances for each size \(N\). 
       $^\dag$ is reported by \citet{bogyrbayeva2023deep}.
    Comp. (\%) represents the relative difference of NICP compared to ICP.
    Times are in seconds.
}
    \label{tab:SetB}
    \vspace{1em}
    \begin{tabular}{>{\raggedright\arraybackslash}m{1.3cm} >{\centering\arraybackslash}m{0.6cm} >{\raggedleft\arraybackslash}m{0.8cm} >{\raggedleft\arraybackslash}m{1.0cm} %
    >{\raggedleft\arraybackslash}m{0.8cm} >{\raggedleft\arraybackslash}m{0.6cm} %
    >{\raggedleft\arraybackslash}m{0.8cm} >{\raggedleft\arraybackslash}m{0.7cm} %
    >{\raggedleft\arraybackslash}m{0.8cm} >{\raggedleft\arraybackslash}m{0.8cm} %
    >{\raggedleft\arraybackslash}m{0.8cm} >{\raggedleft\arraybackslash}m{0.6cm} 
    >{\raggedleft\arraybackslash}m{0.8cm}  >{\raggedleft\arraybackslash}m{0.6cm} %
    >{\raggedleft\arraybackslash}m{0.8cm}
    >{\raggedleft\arraybackslash}m{0.6cm} %
    >{\raggedleft\arraybackslash}m{0.8cm} >{\raggedleft\arraybackslash}m{0.8cm}}
        \toprule
        & & \multicolumn{2}{c}{TSP-ep-all} & \multicolumn{2}{c}{DPS$_{25}$} & \multicolumn{2}{c}{HM$_{4800}$} & \multicolumn{4}{c}{HGA-TAC$^+$} & \multicolumn{2}{c}{ICP} & \multicolumn{2}{c}{NICP} & \multicolumn{2}{c}{Comp. (\%)} \\
        \cmidrule(lr){3-4}\cmidrule(lr){5-6}\cmidrule(lr){7-8}\cmidrule(lr){9-12}\cmidrule(lr){13-14}\cmidrule(lr){15-16}\cmidrule(lr){17-18}
        Dataset & $N$ & Obj & Time & Obj & Time & Obj & Time & Best & Mean & Worst & Time & Obj & Time & Obj & Time & Obj & Time \\
        \midrule
        \multirow{3}{*}{Random}
& 50 & 397.07 & 1.74 & 404.66 & 0.16 & 396.26 & 3.80 & 392.26 & 399.11 & 406.60 & 3.71 & 394.45 & 0.59 & 395.05 & 0.32 & 0.15 & -46.11 \\
& 100 & 535.72 & 33.67 & 548.23 & 0.42 & 544.58 & 14.13 & 536.31 & 545.35 & 554.66 & 10.90 & 534.20 & 1.45 & 535.11 & 0.86 & 0.17 & -40.47 \\
        \cmidrule(lr){2-18}
& Avg. & 466.39 & 17.70 & 476.45 & 0.29 & 470.42 & 8.96 & 464.29 & 472.23 & 480.63 & 7.30 & 464.32 & 1.02 & 465.08 & 0.59 & 0.16 & -42.10 \\
        \midrule
          Amsterdam & 50 & 3.26 & 1.41 & 3.37 & 0.17 & 3.31$^\dag$ & 1.41$^\dag$ & 3.25 & 3.33 & 3.41 & 3.91 & 3.28 & 0.47 & 3.30 & 0.27 & 0.33 & -43.06 \\
        \bottomrule
    \end{tabular}
\end{table}

\end{landscape}

\end{document}